\documentclass{article}

    \PassOptionsToPackage{numbers, compress}{natbib}

    \usepackage[final]{neurips_2022}

\usepackage[utf8]{inputenc} %
\usepackage[T1]{fontenc}    %
\usepackage{hyperref}       %
\usepackage{url}            %
\usepackage{booktabs}       %
\usepackage{amsfonts}       %
\usepackage{nicefrac}       %
\usepackage{microtype}      %
\usepackage{xcolor}         %
\usepackage{colortbl}

\usepackage{booktabs}       %
\usepackage{amsfonts}       %
\usepackage{nicefrac}       %
\usepackage{microtype}      %
\usepackage{xcolor}         %
\usepackage{paralist}       %
\usepackage{graphicx}
\usepackage{amsmath}
\usepackage[ruled,vlined]{algorithm2e}
\usepackage{caption}
\usepackage{subcaption}
\usepackage{diagbox}
\usepackage{bbding}
\usepackage{booktabs}
\usepackage{pifont}
\usepackage{wrapfig}
\usepackage{lipsum}
\usepackage{capt-of}    %
\usepackage{tabularx}   %
\usepackage{multirow}
\usepackage{makecell}
\usepackage{bbm}
\usepackage{pdfpages}
\usepackage{wrapfig,lipsum,booktabs}

\newcommand{\myparagraph}[1]{\textbf{#1}}

\newcommand{\products}{Y}
\newcommand{\product}{y}
\newcommand{\ptext}{\bar{y}}

\newcommand{\patt}{Y_{\text{att}}}
\newcommand{\popt}{Y_{\text{opt}}}
\newcommand{\pprice}{y_{\text{price}}}

\newcommand{\iatt}{U_{\text{att}}}
\newcommand{\iopt}{U_{\text{opt}}}
\newcommand{\iprice}{u_{\text{price}}}

\definecolor{MyDarkBlue}{rgb}{0,0.08,1}
\definecolor{MyDarkGreen}{rgb}{0.02,0.6,0.02}
\definecolor{MyDarkRed}{rgb}{0.8,0.02,0.02}
\definecolor{MyDarkOrange}{rgb}{0.40,0.2,0.02}
\definecolor{MyPurple}{RGB}{111,0,255}
\definecolor{MyRed}{rgb}{1.0,0.0,0.0}
\definecolor{MyGold}{rgb}{0.75,0.6,0.12}
\definecolor{MyDarkgray}{rgb}{0.66, 0.66, 0.66}

\definecolor{MyYellow}{rgb}{254, 246, 170}
\definecolor{MyBlue}{rgb}{170, 217, 251}

\newcommand{\supp}{{Appendix}}

\newcommand{\benchmark}{WebShop} %
\newcommand{\html}{\texttt{HTML}}
\newcommand{\clean}{\texttt{simple}}
\newcommand{\search}[1]{\texttt{search[}{#1}\texttt{]}}
\newcommand{\click}[1]{\texttt{choose[}{#1}\texttt{]}}
\newcommand{\choice}{\texttt{Choice}}
\definecolor{mellowred}{HTML}{CB4042}
\definecolor{mellowblue}{HTML}{0089A7}

\title{\benchmark{}: Towards Scalable Real-World Web Interaction with Grounded Language Agents}

\author{%
  Shunyu Yao\footnotemark[1] \quad Howard Chen\footnotemark[1] \quad John Yang \quad Karthik Narasimhan\\
  Department of Computer Science, Princeton University\\
  \texttt{\{shunyuy, howardchen, jy1682, karthikn\}@princeton.edu}
}

\begin{document}

\maketitle
\renewcommand{\thefootnote}{\fnsymbol{footnote}}
\footnotetext[1]{Equal contribution. Project site with code, data, and demos: \url{https://webshop-pnlp.github.io}.}

\begin{abstract}
Existing benchmarks for grounding language in interactive environments either lack real-world linguistic elements, or prove difficult to scale up due to substantial human involvement in the collection of data or feedback signals.
To bridge this gap, we develop \benchmark{} -- a simulated e-commerce website environment with $1.18$ million real-world products and $12,087$ crowd-sourced text instructions. Given a text instruction specifying a product requirement, an agent needs to navigate multiple types of webpages and issue diverse actions to find, customize, and purchase an item. \benchmark{} provides several challenges for language grounding including understanding compositional instructions, query (re-)formulation, comprehending and acting on noisy text in webpages, and performing strategic exploration.
We collect over $1,600$ human demonstrations for the task, and train and evaluate a diverse range of agents using reinforcement learning, imitation learning, and pre-trained image and language models.
Our best model achieves a task success rate of $29\%$, which outperforms rule-based  heuristics ($9.6\%$) but is far lower than human expert performance ($59\%$).
We also analyze agent and human trajectories and ablate various model components to provide insights for developing future agents with stronger language understanding and decision making abilities. Finally, we show that agents trained on \benchmark{} exhibit non-trivial \textit{sim-to-real} transfer when evaluated on \url{amazon.com} and {\color{black} \url{ebay.com}}
, indicating the potential value of \benchmark{} in developing practical web-based agents that can operate in the wild.

\end{abstract}
\section{Introduction}
Recent advances in natural language processing (NLP) and reinforcement learning (RL) have brought about several exciting developments in agents that can perform sequential decision making while making use of linguistic context~\cite{luketina2019survey,uc2021survey,zhong2021silg}.
On the other hand, large-scale language models like GPT-3~\cite{brown2020language} and BERT~\cite{devlin2019bert} are excelling at traditional NLP benchmarks such as text classification, information extraction and question answering. While the former set of tasks are limited in their set of linguistic concepts and prove difficult to scale up, the latter tasks usually contain static, non-interactive datasets that lack adequate grounding to extra-linguistic concepts~\cite{bender2020climbing}. In order to make further progress in building \textbf{grounded} language models, we believe there is a need for \textbf{scalable} interactive environments that contain: (1) language elements that reflect rich, real-world usage and are collectible at scale, and (2) task feedback that is well-defined and automatically computable to facilitate interactive learning, without the constant need for expensive feedback from humans. %

 The world wide web (WWW) is a massive open-domain interactive environment that inherently satisfies the first aforementioned requirement through its interconnected set of pages with natural text, images and interactive elements.
By being simultaneously \textbf{scalable, semantic, interactive, dynamic and realistic}, the web is uniquely different from existing environments for autonomous agents like games or 3D navigation. Moreover, 
the web also provides a practical environment to deploy trained agents, with great potential for alleviating human efforts in tedious tasks (e.g.\,buying products, booking appointments). 
While there has been prior work on building web-based tasks, they either lack depth in the transition and action spaces, or prove difficult to scale up. Some benchmarks only contain either a single classification task~\cite{pasupat2018mapping, su2017building, mazumder2020flin} or interactions containing only a handful of different pages in each episode~\cite{shi2017world}. Others propose tasks with longer horizons but are either limited to following hyperlinks for web navigation~\cite{nogueira2016end} or require human-in-the-loop feedback due to the lack of an automated reward function~\cite{nakano2021webgpt}. %

\begin{figure}[t!]
    \centering
    \includegraphics[width=1.0\textwidth]{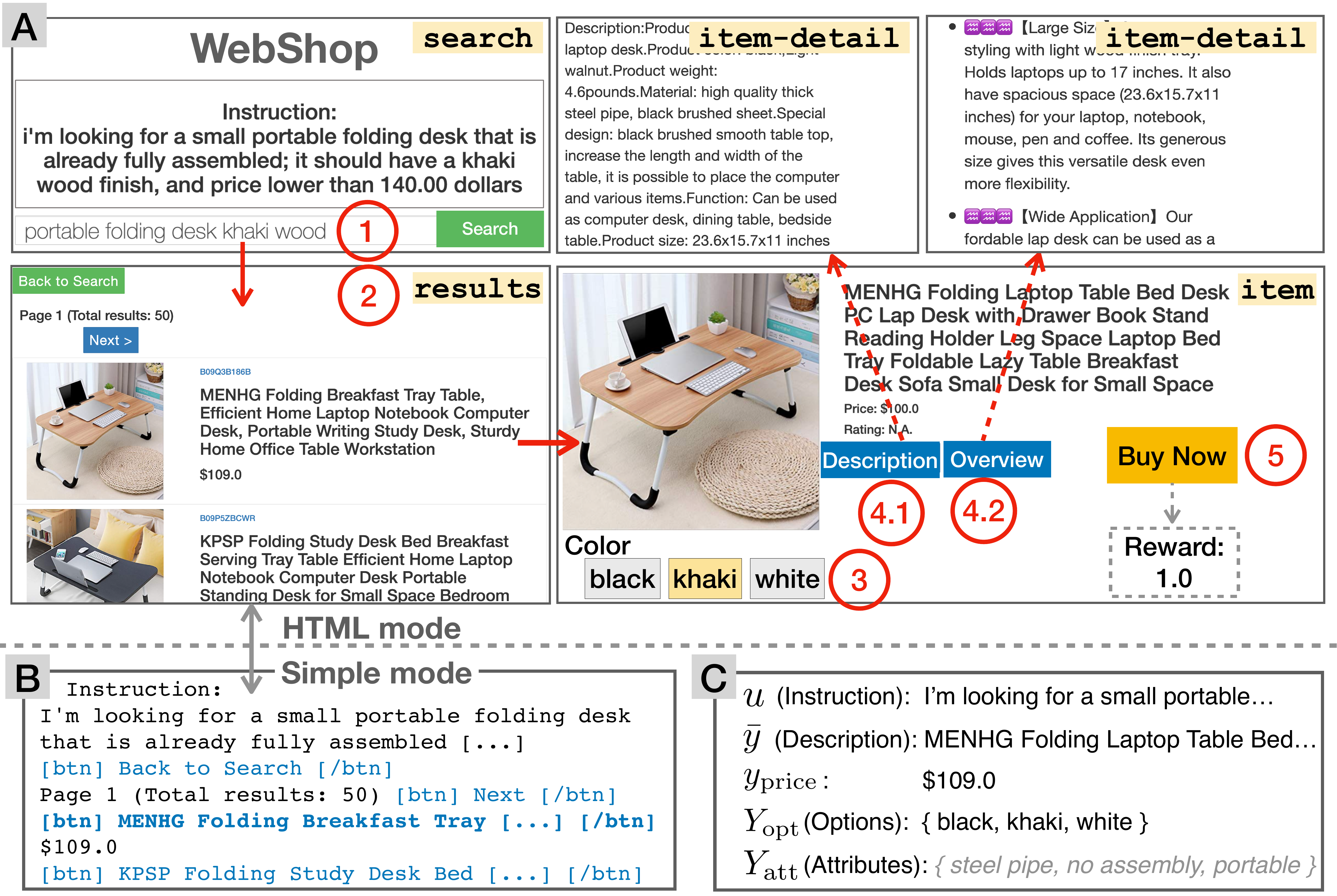}
    \caption{The \benchmark{} environment. \textbf{A}: An example task trajectory in \html{} mode, where a user can (1) search a query in a \texttt{search} page, (2) click a product item in a \texttt{results} page, (3) choose a color option in a \texttt{item} page, (4) check \texttt{item-detail} pages and go back to the \text{item} page, and (5) finally buy the product to end the episode and receive a reward $r \in [0, 1]$ (\S\ref{sec:task}).
    \textbf{B}: the \texttt{results} page in \clean{} mode for agent training and evaluation. The blue text indicates clickable actions and bold text indicates an action selected by the agent.
    \textbf{C}: The product notation used in \S\ref{sec:env} with corresponding examples from the product in \textbf{A}. The attributes $Y_\text{att}$ are hidden from the task performer.
    }
    \label{fig:teaser} 
    \vspace{-20pt}
\end{figure}

In this paper, we introduce \benchmark{} (Figure~\ref{fig:teaser}) -- a large-scale interactive web-based environment for language understanding and decision making -- and train autonomous agents to complete tasks on this benchmark. With the goals of being scalable and containing realistic language and visual elements,
\benchmark{} emulates the task of online shopping on an e-commerce website, where the agent's goal is to understand a human-provided text instruction and \textit{purchase} a product to match the specifications.
To do so, the agent needs to query the website's search engine, choose items to explore from search results, open and read their description and details, and select the necessary options (e.g. 32 oz., red color) before clicking the `Buy' button. 
In order to pick the optimal product that matches user requirements, the agent may need to view and compare various products (including backtracking between pages), and potentially perform multiple searches.
\benchmark{} contains over one million products scraped from \url{amazon.com}, over $12$ thousand crowdsourced instructions, and a diverse semantic action space of searching text queries and choosing text buttons. It is packaged into a convenient OpenAI Gym~\cite{brockman2016openai} environment and can be rendered in two modes (\html{} or \clean{}) with parallel observation spaces that are easy for human and model respectively. Rewards are automatically computed using a combination of programmatic matching functions that consider the attributes, type, options and price of the chosen product, alleviating the need for human evaluation and {\color{black}providing a path to scaling up interactive learning}.

We develop several agents to perform this task, using both reinforcement learning (RL) and imitation learning (IL). We also leverage the latest pre-trained language models~\cite{lewis2019bart,devlin2019bert} for representing and generating text.
Our modular architecture includes a factorized processing of state observations and action choices using ResNets (visual) and Transformers (text), followed by an attention fusion layer that helps the agent contextually score each action. Our best agent achieves an average score of $62.4$ (out of $100$) and successfully completes the task $28.7\%$ of the time, significantly higher than a heuristic baseline that achieves $45.6$ and $9.6\%$, respectively. While this demonstrates the potential for IL and RL, the agents are still much lower than human experts, who can achieve $82.1$ and $59.6\%$ on this task.\footnote{
In our analysis~(\S\ref{sec:breakdown}), we observe that the task requires patience and consistency, which is lacking in some crowdsource workers, leading to lower scores. Even with this caveat, the gap between human performance and the model remains significant. }
We perform several analyses and ablation studies to identify the cause of this gap and find several avenues for agent improvement in the future including more robust search generation, explicit memory modules, and better handling of noisy web text. {\color{black} Finally, we also demonstrate an instance of \textit{sim-to-real} transfer by deploying agents trained with \benchmark{} to operate on \url{amazon.com} and \url{ebay.com}, and find that they can achieve similar performances despite search engine and product differences, and consistently outperform the rule baseline of using the first result returned by the commercial search engines when directly searching the instruction texts.} This demonstrates the practical potential of our work towards developing agents that can operate autonomously on the world wide web (WWW).

\section{Related Work}

\myparagraph{Reinforcement learning on the web.}
\citet{nogueira2016end} introduced WikiNav as a benchmark for RL agents navigating pages, but the task is purely navigational with the actions restricted to either choosing a hyperlink to follow or deciding to stop. 
The World of Bits (WoB) benchmark~\cite{shi2017world} enables training of RL agents to complete tasks on webpages using pixel and Document Object Model (DOM) observations.
Several follow-up papers have tackled MiniWoB using techniques like workflow-guided exploration~\cite{liu2018reinforcement}, curriculum and meta-learning~\cite{gur2018learning}, DOM tree representation~\cite{jia2019dom}, adversarial environment generation~\cite{gur2021adversarial} and large-scale behavioral cloning~\cite{humphreys2022data}. However, MiniWoB lacks long-range decision making across multiple different pages and does not scale easily in terms of difficulty or size due to its use of low-level mouse clicks and keystrokes as actions. In contrast, \benchmark{} requires navigating longer paths with context-based action selection and backtracking, and it uses high-level $search$ and $choose$ actions that are more scalable and transferable to real settings. While not directly operating on web pages, AndroidEnv~\cite{toyama2021androidenv} and MoTIF~\cite{burns2022interactive} provide environments to train agents for interacting with apps and services on mobile platforms. 

\myparagraph{Non-interactive web-based tasks.}
Various supervised classification tasks on webpages have been proposed, including predicting web elements~\cite{pasupat2018mapping}, generating API calls~\cite{su2017building, su2018natural, williams2019automatic} and semantic parsing into concept-level navigation actions~\cite{mazumder2020flin}.  Perhaps most similar content-wise to our work is the Klarna product page dataset~\cite{hotti2021klarna} which contains over $50,000$ product pages labeled with different element categories for supervised classification.
All these works only consider supervised settings with a single decision, and may require the definition of web APIs or command templates for each domain. Our benchmark, \benchmark{}, combines webpages with realistic text and image content with a rich and diverse interaction space for long-range sequential decision making.

\myparagraph{Leveraging the web for traditional NLP tasks.}
Several papers have explored the use of the web for information extraction~\cite{narasimhan2016improving} and retrieval~\cite{adolphs2021boosting}, question answering~\cite{Yuan2020InteractiveMC,Lazaridou2022InternetaugmentedLM}, dialog~\cite{Shuster2022LanguageMT}, and training language models on webtext~\cite{Aghajanyan2021HTLMHP}. These approaches primarily use web search engines  as a knowledge retriever for gathering additional evidence for the task at hand.
Perhaps most similar to our work is WebGPT~\cite{nakano2021webgpt}, which uses a web interface integrated with a search engine to train RL agents to navigate the web and answer questions. However, our environment has a more diverse action and observation space (including images) and does not require human-in-the-loop evaluation.

\section{The \benchmark{} Environment} \label{sec:env}
\vspace{-7pt}
We create \benchmark{} as a large-scale web-based interactive environment with over $1.1$ million real-world products scraped from amazon.com. 
In this environment, an agent needs to find and purchase a product according to specifications provided in a natural language instruction. 
\benchmark{} is designed in a modular fashion which disentangles the website transitions from the task-specific aspects like instructions and reward, allowing for easy extension to new tasks and domains.

\subsection{Task Formulation}
\vspace{-5pt}

\begin{table}[t]
\begin{minipage}{.6\linewidth}
    \centering
\begin{tabular}{lll}
\toprule
     Type &  Argument &  State $\rightarrow$ Next State \\
\midrule
    \texttt{search} & [\textit{Query}] & Search $\rightarrow$ Results\\ 
    \texttt{choose} & Back to search & $*$ $\rightarrow$ Search\\
    \texttt{choose} & Prev/Next page & Results $\rightarrow$ Results\\
    \texttt{choose} & [\textit{Product title}] & Results $\rightarrow$ Item\\
    \texttt{choose} & [\textit{Option}] & Item $\rightarrow$ Item\\
    \texttt{choose} & Desc/Overview & Item $\rightarrow$ Item-Detail\\
    \texttt{choose} & Previous & Item-Detail $\rightarrow$ Item\\
    \texttt{choose} & Buy & Item $\rightarrow$ Episode End\\
\bottomrule
\end{tabular}
\vspace{5pt}
\caption{Actions in \benchmark.}
\label{table:action_space_}
\end{minipage}%
\hspace{5pt}
\begin{minipage}{.39\linewidth}
    \centering
\includegraphics[width=.9\textwidth]{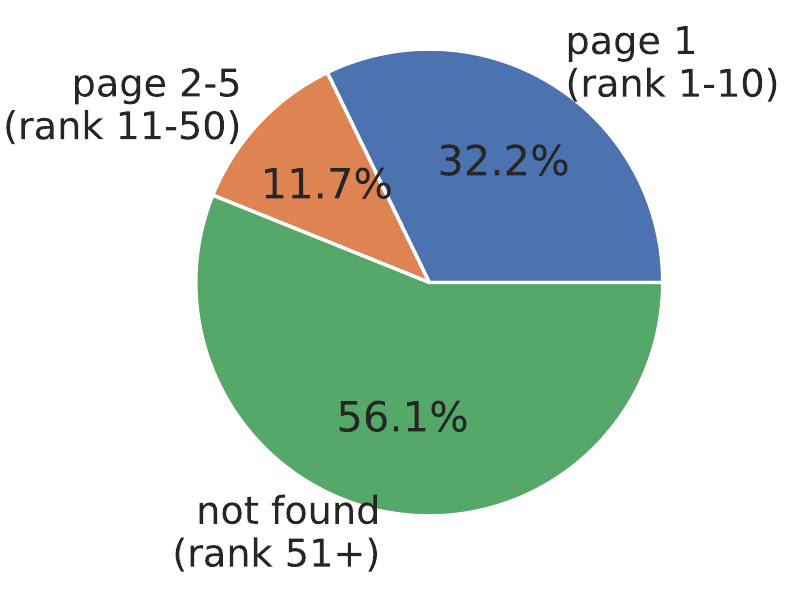}
\captionof{figure}{\color{black}Item rank in search results when the instruction is directly used as search query.}
\label{fig:search}
\end{minipage}%
\vspace{-15pt}
\end{table}

\benchmark{} can be formulated as a partially observable Markov decision process (POMDP) $(\mathcal{S}, \mathcal{A}, \mathcal{T}, \mathcal{R}, \mathcal{U}, \mathcal{O})$ with state space $\mathcal{S}$, action space $\mathcal{A}$, deterministic transition function $\mathcal{T}: \mathcal{S} \times \mathcal{A} \to \mathcal{S}$, reward function $\mathcal{R}: \mathcal{S} \times \mathcal{A} \to [0, 1]$, instruction space $\mathcal{U}$, and a state observation space $\mathcal{O}$.

\myparagraph{State and action.} A state $s \in \mathcal{S}$ represents a web page, which falls into one of the four types -- the \emph{search} page that contains a search bar, the \emph{results} page that lists a set of products returned by a search engine, the \emph{item} page that describes a product, or the \emph{item-detail} page that shows further information about the product (Figure~\ref{fig:teaser}A(1-4) respectively).
We define the following notations for a product $\product$. We denote $\bar{\product}$ to be the aggregation of the various text fields including product title, description, and overview. We denote $\pprice$ to be the price, $\products_{\text{opt}}$ to be a set of buying options, and $I$ to be a set of images, each corresponding to a specific option. Finally, each product is associated with $\products_{\text{att}}$, a set of attributes hidden from the agent which is extracted from the title and the \textit{item-detail} pages (\S\ref{sec:task}). The attributes are used for the automatic reward calculation.

An action $a \in \mathcal{A}(s)$ can either be searching a text query (e.g.\,$\search{\text{Red shoes}}$) or choosing a text button (e.g.\,$\click{\text{Size 9}}$) as shown in Table~\ref{table:action_space_}. 
These two action types are not available simultaneously -- search is only allowed when the agent is at the search page; on all other pages, click is the only action choice.
The chosen action argument (button) will be clicked as a web link as opposed to the low-level mouse-click actions in previous environments such as World of Bits~\cite{shi2017world}. 
The transitions initiated by clicks deterministically redirect the web page to one of the four page types (Table~\ref{table:action_space_}). The transition initiated by search is based on a deterministic search engine (\S\ref{sec:task}).

\myparagraph{Observation.} Using Flask~\cite{flask} and OpenAI Gym~\cite{brockman2016openai}, we provide two parallel observation modes to render the state and instruction $\mathcal{S} \times \mathcal{I} \to \mathcal{O}$: (1) \html{} mode that contains the HTML of the web page, allowing for interaction in a web browser(Figure~\ref{fig:teaser}A), and (2) \clean{} mode which strips away extraneous meta-data from raw HTML into a simpler format (Figure~\ref{fig:teaser}B). The human performance scores in \S\ref{sec:methods_il} are collected in the \html{} mode, while all models are trained and evaluated in the \clean{} mode. Note that while the environment allows for training reinforcement learning agents on raw pixels in \html{} mode (like in \citet{shi2017world}), we believe that it provides a very low-level non-semantic action space. Moreover, it is straightforward to write a translator that converts any new HTML page into \clean{} format for use with trained agents, which enables sim-to-real transfer.

\myparagraph{Instruction and reward.}
Each natural language instruction $u \in \mathcal{U}$ contains the following information: a non-empty set of attributes $\iatt$, a set of options $\iopt$, and a price $\iprice$. The instruction is generated based on a target product $y^*$ by human annotators. The instruction collection process is lightweight and scalable (\S\ref{sec:task}).
Concretely, $\iatt \subseteq \patt^*$ is a subset of the product attributes,
$\iopt \subseteq \popt^*$ is a subset of the product option field-value pairs,
$\iprice > \pprice^*$ is a price set to be higher than the target product price.
For example, the instruction ``Can you find me a pair of \textit{black-and-blue} sneaker that is \textit{good in rain weather}? I want it to have \textit{puffy soles}, and price less than $90$ dollars.'' contains the aforementioned attributes $\iatt = \{\text{``waterproof''}, \text{``soft sole''}\}$ and option $\iopt = \{ \text{``color'': ``black and blue''} \}$.
In each episode, the agent receives a reward $r = \mathcal{R}(s_{T}, a)$ in the end at timestep $T$, where $a = \click{\text{buy}}$, $y$ is the product chosen by the agent in the final state $s_{T}$, and $\patt$ and $\popt$ are its corresponding attributes and options. The reward is defined as:%
\begin{equation}
\label{eq:r}
    r = r_{\text{type}} \cdot \frac{|\iatt \cap \patt| + |\iopt \cap \popt| + \mathbf{1}[\pprice \le \iprice]}{|\iatt| + |\iopt| + 1}
\end{equation}
where the type reward $r_{\text{type}} = \texttt{TextMatch}(\ptext, \ptext^*)$ is based on text matching heuristics to assign low reward 
when $\product$ and $\product^*$ have similar attributes and options but are obviously different types of products. For example, ``butter'' and ``plant-based meat'' differ in types but may both contain attributes ``cruelty-free'', ``non-GMO'', and an option ``size: pack of 2''. The exact formula for $\texttt{TextMatch}(\cdot)$ is in the \supp~\S\ref{rewardCalc}. 

{\color{black}
\myparagraph{Evaluation metrics.}
We use two evaluation metrics: (1) \textbf{Task Score}: defined as $(100 \times \text{avg. reward})$, which captures the average reward obtained across episodes; and (2) \textbf{Success Rate (SR)} defined as the portion of instructions where $r=1$. Note that it is possible to obtain $r=1$ for an episode even if the final product is not $y^*$ --- for example, there could be many items that satisfy the goal ``I want a red shirt'', even if the goal is generated from a specific red shirt item.}

\subsection{Environment Implementation} \label{sec:task}

\myparagraph{Data scraping.} We use ScraperAPI \cite{scraperapi} to scrape $1,181,436$ products from \url{amazon.com} across $5$ categories (fashion, makeup, electronics, furniture, and food) using $113$ sub-category names as queries. The product texts (title and item details) have an average length of $262.9$ and a vocabulary size $224,041$ (word frequency higher than $10$). In addition, the products have a total of $842,849$ unique options, reflecting the scale and complexity of the data.
More details about product scraping is in the \supp~\S\ref{appx:scrape}.

\myparagraph{Search engine.} We use Pyserini \cite{lin2021pyserini} for the search engine, where indices are built offline using a BM25 sparse retriever with text for each product concatenated from the title, description, overview, and customization options. The search engine is deterministic, which eases imitation learning and result reproducibility. More details in \ref{appx:search_engine}.

\myparagraph{Attribute mining and annotation.}
Each product is annotated with a set of hidden \textit{attributes}, which are used to represent its latent characteristics as well as to calculate the reward as detailed in \S\ref{sec:env}. An attribute is a short natural language phrase that describes the property of the product (see examples in Figure~\ref{fig:teaser}). 
We mine the attributes by calculating TF-IDF scores for all bi-grams in the concatenated titles and descriptions based on each product category.
We review the top $200$ bi-grams for each category, remove the noisy ones by inspection (decide based on whether the bi-gram is human understandable), and assign them to the products. We consolidate a pool of $670$ attributes. 
See more details in the \supp~\S\ref{appx:attribute}.

\myparagraph{Natural language instructions.}
 We use Amazon Mechanical Turk (AMT) to collect natural language instructions that specify goal products with appropriate options. Specifically, an AMT worker is presented with a sampled goal product, including the product title, category, attributes, and the buying options, and asked to write a command to instruct an automatic shopping agent to find the target. Workers are instructed to avoid being too specific such as including the entire title in the instruction, but stay faithful to describing the target product. We collect a total of $12,087$ linguistically diverse instructions with an overall vocabulary size of $9,036$ words and an average length of $15.9$ words. We provide the detailed annotation process and interface in the \supp~\S\ref{appx:instruction}.

{\color{black}
\myparagraph{Human demonstrations.} 
We collect trajectories from humans performing the task in the \html{} mode of \benchmark{} to understand the task difficulty for humans and to analyze how humans would solve the task.
We use qualification tests to train and select motivated workers to perform the task.
We recruit and train a total of $13$ workers for data collection, and among them we select the top $7$ performing workers to be ``experts'' (see \supp~\S\ref{appx:traj} for examples). We also leverage this data to perform imitation learning (described in \S\ref{sec:methods_il}).
}

{\color{black}

\subsection{Research Challenges}
\label{sec:challenge}
\benchmark{} brings together several research challenges for autonomous systems from various subfields in NLP and RL into a single benchmark. These include: 1) generation of good search queries~\cite{komeili2021internet,zhuang2022bridging} and reformulation~\cite{nogueira2017task,wang2020deep}, 2) strategic exploration for navigating through the website~\cite{Yao2020KeepCA,yao2021reading,liu2018reinforcement}, 3) robust language understanding for textual state and action spaces~\cite{andreas2020task,budzianowski2018multiwoz,hausknecht2020interactive,shridhar2020alfred}, and 4) long-term memory for comparing items or backtracking~\cite{wayne2018unsupervised,fortunato2019generalization,lampinen2021towards} (Figure~\ref{fig:teaser}). While we believe individual advances in each of these will improve agent performance, \benchmark{} also provides an ideal testbed for the development of interdisciplinary techniques that tackle more than one of the above mentioned challenges simultaneously. For example, external memory modules may be very effective if combined with strategic exploration, or exploration could be helpful in information query reformulation. Further analysis based on human and model trajectories is in \S\ref{sec:breakdown}.

}

\section{Methods}
\vspace{-7pt}

\begin{figure}[t!]
    \centering
    \includegraphics[width=.94\textwidth]{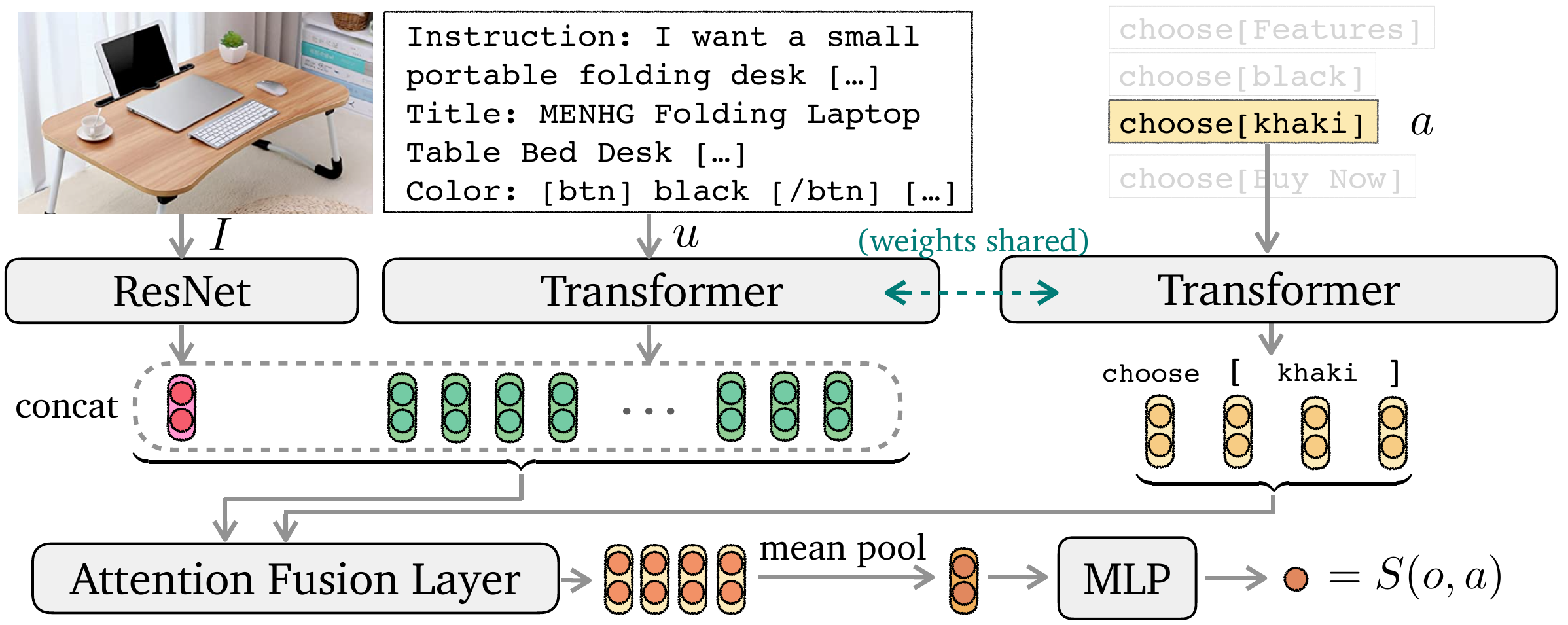}
    \caption{Architecture of our choice-based imitation learning (IL) model. The image $I$ is passed to a ResNet to obtain the image representation. The instruction text $u$ is passed to a transformer (initialized with BERT) to obtain the text representations. The concatenated bi-modal representations are fused with the action representations using the Attention Fusion Layer. The resulting fused-action representations are mean-pooled and reduced by an MLP layer to a scalar value $S(o, a)$ denoting the logit value of the action $\click{\text{khaki}}$.}
    \label{fig:model} 
    \vspace{-8pt}
\end{figure}
We propose various models that combine language and image pre-training with imitation learning (IL) and reinforcement learning (RL). More details are provided in the \supp~\S\ref{appx:model}.

\subsection{Rule Baseline}
\vspace{-5pt}
A simple \textbf{rule baseline} is to search the exact instruction text, then choose and buy the first item in the results page without choosing any options. 
The heavy lifting of the lexical search engine makes it also a simple non-learnable information retrieval (IR) baseline, and would lead to a non-trivial attribute reward.
However, simple heuristic rules cannot resolve noisy natural language options, strategically explore, or learn to generate what to search, so the total reward and task success rate should be low. 

\subsection{Imitation Learning (IL)}
\label{sec:methods_il}
\vspace{-5pt}
For the text generation and choice problems presented in \benchmark{}, we propose using two pre-trained language models to separately learn how to search and choose from human demonstrations.

\myparagraph{Imitating human search generation.} We frame searching as a sequence-to-sequence text-generation problem: the agent generates a search action $a = \search{\dots}$ given an instruction $u$ without considering any other context (e.g.\,past searches, visited items). We use $M=1,421$ instruction-search pairs from $1,012$ training human trajectories to construct a dataset $\mathcal{D} = \{(u, a)\}_{i=1}^{M}$ and fine-tune a BART model~\cite{lewis2019bart} parameterized by $\phi$ to perform conditional language modeling:
\begin{equation}
    \mathcal{L}_{\text{search}} = \mathbb{E}_{u, a \sim \mathcal{D}}\left[-\log \pi_{\phi}(a\mid u)\right]
\end{equation}
\myparagraph{Imitating human choice.} The choice-based imitation model (Figure~\ref{fig:model}) predicts a probability distribution over all the available click actions $\mathcal{A}(o)$ in observation $o$ and maximizes the likelihood of the human clicked button $a^* \in \mathcal{A}(o)$. We construct a dataset $\mathcal{D}' = \{(o, \mathcal{A}(o), a^*)\}_{i=1}^{M'}$ of  $M' = 9,558$ samples from the training human trajectories.
We use a $12$-layer pre-trained BERT model~\cite{Devlin2019BERTPO} parameterized by $\theta$ to encode the $o$ into an observation representation of contextualized token embeddings, and we similarly encode each action. 
Each action representation is passed into a cross-attention layer with the observation representation, then mean pooled into a single vector and multiplied with a matrix $W$ to obtain a scalar score $S(o, a)$. The policy $\pi_\theta \left(a \mid o, \mathcal{A}(o) \right)$ is the softmax distribution over action scores $S(o, a)$:
\begin{align}
    \mathcal{L}_{\text{choose}} &= \mathbb{E}_{o, \mathcal{A}(o), a^* \sim \mathcal{D}'} \left[ - \log \pi_\theta \left(a^* \mid o, \mathcal{A}(o) \right) \right]\\
    \pi_\theta \left(a \mid o, \mathcal{A}(o) \right)  &\sim \exp \left( W^\top \text{mean} \big[ \text{cross-attn} \big(\text{BERT}(o; \theta), \text{BERT}(a; \theta) \big) \big] \right) 
\end{align}
\myparagraph{Handling Images.}
We use a pre-trained ResNet-50~\cite{He2016DeepRL} to pre-process images across different products and options into a $512$ dimensional feature vector, which is then transformed into $768$ dimensions with a learned linear layer and concatenated to $\text{BERT}(o)$ as the observation representation.

\myparagraph{Full pipeline.} Combining the above during environment interaction, we use the BART model in the search page to generate the top-$5$ search queries via beam search and choose a random one.
For other pages, we sample one action from $\pi_\theta \left(a \mid o, \mathcal{A}(o) \right)$ using the BERT model. 
We find these methods useful to encourage diverse actions. In contrast, an ineffective strategy that uses only the top generated search query or the button with the highest probability might lead to limited product candidates or being stuck (e.g.\,bouncing back and forth between pages). 

\subsection{Reinforcement Learning (RL)}
We also fine-tune the choice-based IL model with online RL (i.e.\,IL+RL). Prior work suggests that directly fine-tuning text generation via RL might lead to language drifting~\cite{Lazaridou2020MultiagentCM} and deteriorated performance. Therefore, we freeze the BART model to provide the top-$10$ search generations as a refined action space for the choice-based IL model to learn to pick -- an inspiration borrowed from previous work in text games~\cite{Yao2020KeepCA} and referential games~\cite{Lazaridou2020MultiagentCM}. 
We use the policy gradient method~\cite{mnih2016asynchronous} with return-to-go $R_t = \mathbb{E}_{\pi}[r_t + \gamma R_{t+1}]$ and a learned value baseline $V(o) = W_v^\top \text{BERT}(o; \theta)$ parameterized by $\{W_v, \theta \}$ (the BERT weights are tied with the policy):
\begin{equation}
    \mathcal{L}_{\text{PG}} = \mathbb{E}_{\pi} \left[ - \left(R_t - V(o_t)\right) \log \pi \left(a_t \mid o_t, \mathcal{A}(o_t) \right) \right]
\end{equation}
The value $V(o)$ is learned with an L2 loss $\mathcal{L}_{\text{value}} = (R_t - V(o_t))^2$. We also add an entropy loss $\mathcal{L}_{\text{entropy}} = \sum_{a \in \mathcal{A}(o_t)} \pi_\theta \big(a_t \mid o_t, \mathcal{A}(o_t) \big) \log \pi_\theta \big(a_t \mid o_t, \mathcal{A}(o_t) \big)$ to prevent premature convergence. Our full RL model minimizes the total loss $\mathcal{L}_{\text{RL}} = \mathcal{L}_{\text{PG}} + \mathcal{L}_{\text{value}} + \mathcal{L}_{\text{entropy}}$.
\vspace{-8pt}

\section{Experiments}
\vspace{-4pt}
\subsection{Setup and task verification}
\vspace{-5pt}
We split a total of $12,087$ instructions into an i.i.d.\,distributed train / development / test split of $10,587$ / $1,000$ / $500$ instances for all models. {While future work can investigate splits with more generalization gaps (e.g.\,split by product category), we will show the i.i.d.\,split is already challenging for current models.} We randomly sample a subset of the $10,587$ training instructions, then collect $1,012$ human demonstrations for task verification and imitation learning (IL) and a further $54$ demonstrations from instances in the development set for IL hyperparameter tuning and checkpoint selection. We also collect human trajectories for all $500$ test instructions and report human and model performances averaged across these $500$ instructions.
More setup details are in the \supp~\S\ref{appx:exp_setup}.

\begin{figure}[t]
    \centering
    \includegraphics[width=1.0\textwidth]{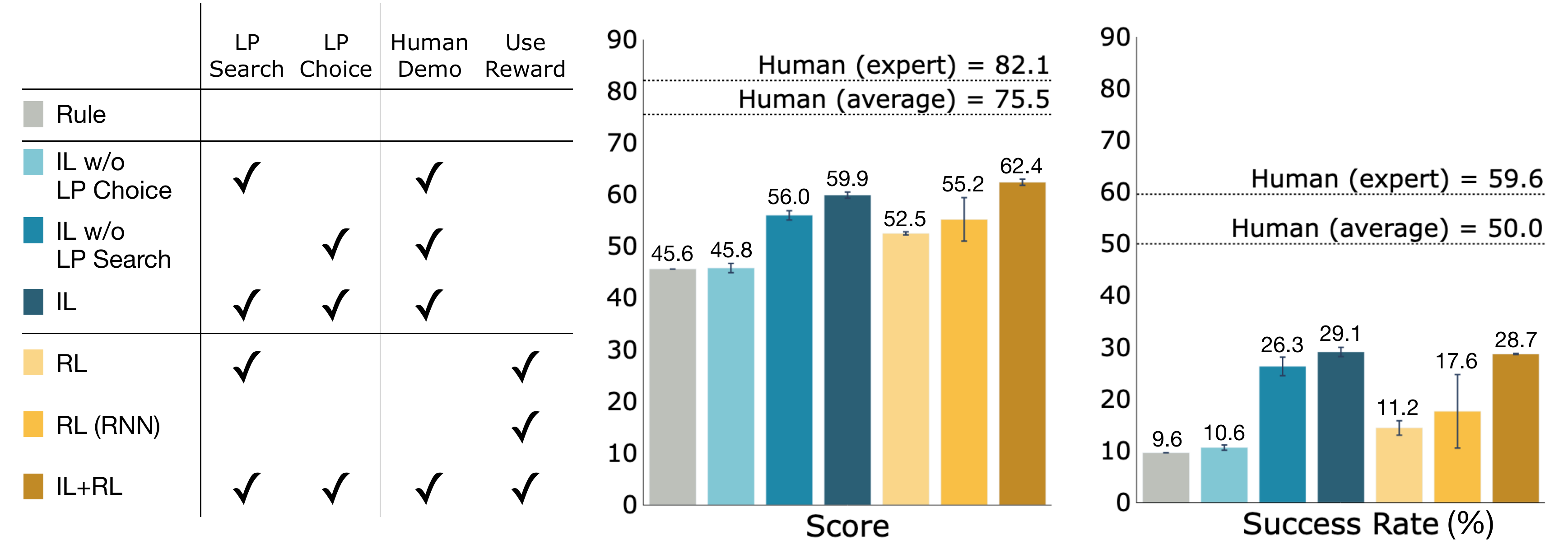}
    \caption{
        Task scores and Success Rate (\%) for our models on the test split of \benchmark{} over $3$ trials. LP Search uses a pre-trained BART model to generate the search query and IL w/o LP Search uses the rule-based heuristic. LP Choice uses pre-trained BERT weights to initialize the choice action model and IL w/o LP Choice trains a Transformer from scratch.
    }
    \label{fig:main_results} 
    \vspace{-3pt}
\end{figure}

\subsection{Results}
\label{sec:results}
\vspace{-5pt}
\myparagraph{Task performance.} From Figure~\ref{fig:main_results}, we observe that the rule baseline obtains a low score of $45.6$ and a very low success rate of $10\%$ since it cannot resolve options specified in language or explore more products, empirically demonstrating the non-trivial nature of the task. The IL model significantly outperforms the rule baseline on both metrics, achieving a score of $59.9$. Further RL finetuning improves the score to $62.4$ while slightly hurting the success rate ($29.1\% \rightarrow 28.7\%$) (analyzed further in \S\ref{sec:breakdown}).
We also observe a significant gap between models and humans -- our best model's success rate ($29.1\%$) is less than half of expert humans ($59.6\%$) and only $60\%$ of the average human ($50\%$). This indicates a great room for model improvement by tackling reseach challenges in \benchmark{}. 

\myparagraph{IL ablations.}
Figure~\ref{fig:main_results} also contains several ablations that confirm important design choices for models. When the choice action model for the IL agent is randomly initialized (\textbf{IL (w/o LP Choice)}; {\color{black} LP = language-pretraining}), the success rate drops by nearly two-thirds, indicating the importance of language pre-training for our task.
When the search query generator in the IL agent is replaced by a simple rule, which always uses the instruction text (\textbf{IL (w/o LP Search)}), both reward and success rate drop by around $3$ points. This suggests the importance to explore by expanding the search space for exploration, but it is not as critical as learning to choose the right options.
We experiment with incorporating history of one past observation and the last five actions into the model and find a slight degradation in the score from $59.9$ to $57.3$, suggesting more advanced techniques are needed to leverage past information.
More ablations in \S\ref{appx:exp_setup}.

\myparagraph{RL ablations.} 
When we directly train an RL agent (\textbf{\text{RL}}) from pre-trained BERT parameters, the performance is even worse than the rule baseline. This suggests that IL warm-starting is critical, possibly because of the significant domain shift from traditional language tasks. We also consider a simple RL model with RNN text encoders instead of the Transformer (\textbf{RL (RNN)}), which has a success rate more than $10\%$ 
worse than the IL + RL model with a much larger variance. 
We hypothesize that RL with a more powerful architecture could help boost and stabilize the performance if the model is initialized with better language and task priors.

\subsection{Analysis}
\label{sec:breakdown}
\vspace{-4pt}

To better understand the differences between the agents and human experts, we perform several fine-grained analyses. We first break down the overall score into its four sub-parts according to Eq.~\eqref{eq:r}: 1) attribute score ($|\iatt \cap \patt| / |\iatt|$), 2) option score ($|\iopt \cap \popt| / |\iopt|$), 3) price score ($\mathbf{1}[\pprice \le \iprice]$), and 4) type score ($r_{\text{type}}$). We  report trajectory statistics such as the average number of states, unique items visited, and number of searches per episode in Table~\ref{table:breakdown} and provide qualitative examples of the trajectories in Table~\ref{table:example}.

\begin{table}[t]
    \centering
    \footnotesize
\begin{tabular}{llllll|lll}
\toprule
    & \multicolumn{5}{c}{Score} & \multicolumn{3}{c}{Count}\\
    & All & Att & Opt & Type & Price & State & Item & Search\\
\midrule
    Rule           & 45.6 & 66.6 & 0.0 & 80.5 & 86.0 & 3.0 ~~~~~~(3 / 3) & 1.0 ~~(1 / 1) & 1.0 ~~(1 / 1)\\
    IL             & 59.9 & 69.3 &  \textbf{45.2} & 86.4 & 84.0 & 9.4 ~~~~(90 / 3) & 1.6 (11 / 1) & 1.3 (17 / 1)\\
    IL+RL        & \textbf{62.4} & \textbf{74.0} & 38.9 & \textbf{89.7} & \textbf{88.7} & 4.5 ~~~~~~(5 / 1) & 1.0 ~~(1 / 1) & 1.0 (~~1 / 1)\\ 
    \midrule
    Human Expert & 82.1 &  81.8 & 73.9 & 94.4 & 97.7 & 11.3 (114 / 4) & 1.9 (16 / 1) & 1.4 (16 / 1)\\
\bottomrule
\end{tabular}%
\vspace{3pt}
\caption{
    Left: Score breakdown. Right: average, maximum, and minimum number of states visited, items checks, and searches in a trajectory.
}
\label{table:breakdown}
\vspace{-10pt}
\end{table}

\begin{table}[t]
\centering
\footnotesize

\begin{minipage}{.49\linewidth}
\begin{tabularx}{\textwidth}{X}
\toprule

{\centering\textbf{Instruction 1}}\\
\midrule
{I want to find \textcolor{mellowred}{white blackout} shades that are     \textcolor{mellowred}{66 inches in width and 66 inches in height}. They need to be \textcolor{mellowblue}{easy to install} [...]}
\\
\midrule
\textbf{Human Actions } ($r=1.0$, $\text{length}=8$)\\
\texttt{search[} 66 inches in width and 66 inches in height white shades \texttt{]} \\
\click{ item : CALYX... } \\
\click{ Back to Search } \\
\search{ 66 x 66 blackout shades } \\
\click{ item : Milin... } \\
\click{ opt : \textcolor{mellowred}{66"w x 66"h~}} \\
\click{ opt : \textcolor{mellowred}{cordless bottom up-blackout-white~}} \\
\click{ Buy } \\
\\
\\
\midrule
\textbf{IL+RL Actions } ($r=0.2$, $\text{length}=3$) \\
\texttt{search[} white blackout shades 65 inches in width and 66 inches in height \texttt{]} \\
\click{ item : Window... } \\
\click{ Buy } \\
\bottomrule
\end{tabularx}

\end{minipage}
\begin{minipage}{.49\linewidth}

\begin{tabularx}{\textwidth}{X}

\toprule
\textbf{Instruction 2} \\
\midrule 
{I need a \textcolor{mellowred}{gingko light} and \textcolor{mellowred}{20"x20"} pillow cover that is \textcolor{mellowblue}{hand painted} [...]
}
\\
\\
\midrule 
\textbf{Human Actions} ($r=1.0$, $\text{length}=17$) \\
\texttt{search[} gingko light 20"x20" pillow cover  \textcolor{mellowblue}{hand painted} \texttt{]} \\
\click{ item : Maison... } 
$[$...$]$ \\
\click{ Description } \\
\click{ < Previous } \\
\click{ Overview } \\
\click{ < Previous } 
$[$...$]$ \\
\click{ item : Maison... } \\
\click{ opt : \textcolor{mellowred}{20"x20"} } \\
\click{ opt : \textcolor{mellowred}{nudes (gingko light) }} \\
\click{ Buy } \\
\midrule
\textbf{IL+RL Actions} ($r=0.25$, $\text{length}=3$) \\
\texttt{search[} gingko light and 20x20 pillow cover \textcolor{mellowblue}{hand painted} \texttt{]} \\
\click{ item : UPOOS... } \\
\click{ Buy } \\
\bottomrule
\end{tabularx}
\end{minipage}
\vspace{5pt}
\caption{Two example trajectories (showing only actions) from the human and the IL+RL model. We omit some human actions from instruction 2 for space and truncate the item names for readability. \textcolor{mellowred}{Red} denotes options and \textcolor{mellowblue}{blue} denotes attributes.
}
\label{table:example}
\vspace{-4pt}
\end{table}

\myparagraph{Human expert vs.\,agents.}
Human experts outperform the agents on all score sub-parts (Table~\ref{table:breakdown}), but the most significant boost comes from the option score (a $28\%$ gap), revealing that agents have trouble selecting the correct product options.
Humans also have longer trajectories, explore more items and perform more searches than the agents, with a higher variance, demonstrating their flexibility. 
Table~\ref{table:example} provides some samples trajectories.
In the first example, the human decides to search again after removing `inches', `width', `height', and `white' from the query since product texts often contain abbreviated symbols for these terms like `"', `w', and `h'. Thus, \textbf{search generation} is challenging for models since it involves reasoning and adapting to grounded environments, and ideas from query reformulation~\cite{nogueira2017task,adolphs2021boosting} could help alleviate this. Agents also struggle to perform robust \textbf{semantic matching}, which is important in choosing options that contain noisy paraphrases of instruction spans. In the second example, the human explores several products first, and decides to return to the first explored product, demonstrating long-term \textbf{memory} that is lacking in the IL+RL model. 

\begin{table}[t]
    \centering
 \resizebox{.8\linewidth}{!}{%
\begin{tabular}{rcccc}%
\toprule
    & Instr. text & IL BART & Human expert (first) & Human expert (last)\\ 
    \midrule
Score & 94.9 & 94.5 & 94.5 & \textbf{95.5} \\
Success Rate & 85.4\% & 84.2\% & 85.6\% & \textbf{87.8\%} \\
\bottomrule
\end{tabular}%
}
\vspace{4pt}
\caption{
    \color{black}{Task performance with the \choice{} oracle. \textit{first} and \textit{last} refer to the first and last search queries found in human demonstrations, respectively.}
}
\label{table:oracle}
\vspace{-13pt}
\end{table}

\myparagraph{Effect of RL fine-tuning after IL.} Table~\ref{table:breakdown} also shows that RL fine-tuning adapts the IL model to become more `greedy' and less `exploratory', as the average trajectory length drops from $9.4$ to $4.8$, and the model explores fewer items and search queries. As a result, the attribute, type, and price scores all increase, but option score drops from $45.2$ to $38.9$. This points to the need for a better balance exploration with exploitation during RL, e.g.\,by using intrinsic bonuses.

{\color{black}
\paragraph{Results with at \choice{} oracle.} 
To disentangle the effects of learning to search from choosing the right actions, we construct a \choice{} oracle that has access to the hidden reward function as well as hidden attributes and options underlying each product and instruction.\footnote{\color{black}A similar search oracle is also possible but harder to design since the search space is infinite. One possible oracle is to search for the underlying product name for each instruction, but that makes choice trivial as the underlying product is then almost always the first search result.} Given a search query, the \choice{} oracle will perform an exhaustive search over every result item, try out all combinations of options and finally choose the best item with options that maximize the reward --- meaning each episode will take hundreds or thousands of steps, as opposed to $4.5$ and $11.3$ steps on average for the IL+RL model and human experts (Table~\ref{table:breakdown}).
We use 500 test instructions and consider four types of search queries: the instruction text (used by rule baseline), top IL BART generated query (used by all learning models), and the first and last queries from human experts in each test trajectory.\footnote{$74.8\%$ of the time there is only one query in the trajectory.} \choice{} oracle improves the success rate of rule heuristics from $9.6\%$ to $85.4\%$, and even the human expert success rate from $59.6\%$ to $87.8\%$ (Table~\ref{table:oracle}), confirming that choosing the right actions is indeed a major bottleneck for current models with great room for improvement. However, using a better search query is still important even with such a strong \choice{} oracle, as the last human search query still outperforms other search queries. This also suggests human experts improve search query qualities over reformulations.}

\subsection{Zero-shot Sim-to-real Transfer}
\label{sec:sim2real}
\vspace{-5pt}
Finally, we conduct a `\textit{sim-to-real}' transfer experiment where our models trained on \benchmark{} are tested on the real-world Amazon (\url{amazon.com}) and {\color{black}eBay (\url{ebay.com})} shopping websites without any fine-tuning.
We sample $100$ test instructions and deploy $3$ \benchmark{} models (rule, IL, IL+RL) to interact with Amazon and eBay, and manually score each episode based on Eq.~\eqref{eq:r}.
As shown in Table~\ref{table:sim_real_stats}, model performances on the two website are similar to \benchmark{} performances in Figure~\ref{fig:main_results}, except for the rule baseline, likely due to the better search engine of Amazon than \benchmark{}.

\vspace{-5pt}

\begin{table}[h]
    \centering
    \footnotesize
\begin{tabular}{cccccc|cccccc}
\toprule
    & \multicolumn{5}{c}{Amazon} & \multicolumn{5}{c}{eBay}\\
    & Score / SR & Att & Opt & Type & Price & Score / SR & Att & Opt & Type & Price\\
\midrule
    Rule  & 45.8 / 19\% & 45.6 & 38.0 & 66.2 & 90.0  & 31.7 / ~~7\% & 62.3 & 25.9 & 49.0 & 67.0 \\
    IL    & 61.5 / 27\% & 60.7 & \textbf{53.7} & 85.6 & 96.0  & 58.2 / \textbf{21\%} & 60.2 & \textbf{52.3} & 85.1 & 96.9 \\
    IL+RL & \textbf{65.9} / \textbf{25\%} & \textbf{71.6} & 47.0 & \textbf{87.8} & \textbf{100.0} & \textbf{62.3} / \textbf{21\%} & \textbf{69.1} & 39.5 & \textbf{91.7} & \textbf{97.0} \\
\midrule
    Human & 88.2 / 65\% & 86.2 & 76.3 & 99.0 & 100.0 & 79.7 / 40\% & 80.3 & 70.1 & 99.5 & 100.0 \\
\bottomrule
\end{tabular}
\vspace{8pt}
\caption{Zero-shot sim-to-real transfer to Amazon and eBay over $100$ test instructions. The Score / SR (Success Rate) column indicates the overall performance. The remaining breakdown are in Score.}
\label{table:sim_real_stats}
\vspace{-13pt}
\end{table}

On \url{amazon.com}, IL+RL achieves a Score of $65.9$ and SR of $25\%$, outperforming the Rule baseline's Score of $45.8$ and SR of $19\%$ by large margin. Similarly, on \url{ebay.com}, IL+RL achieves a Score of $62.3$ and SR of $21\%$, widely outperforming the Rule baseline's Score of $31.7$ and SR of $7\%$.
These results confirm positive sim-to-real values of trained agents for real-world web tasks despite domain shifts in data (products) and dynamics (search engine).
We also obtain a human average score of $88.0$ / $79.7$ and success rate of $65\%$ / $40\%$ by asking turkers (\S\ref{sec:task}) to find the instructed product on the Amazon and eBay websites respectively. While humans perform much better than agents, their web interactions are much slower --- taking on average $815$ seconds per episode as opposed to $< 8$ seconds per episode for our IL and IL+RL models on Amazon.
This sim-to-real transfer only requires two minor coding additions, suggesting that environments like \benchmark{} are suitable for developing \textit{practical} grounded agents to reduce human effort on real-world web tasks.
We provide additional performance and in-depth analysis in \supp~\S\ref{appx:sim2real}.
\section{Discussion}
\vspace{-7pt}
We have developed \benchmark{}, a new web-based benchmark for sequential decision making and language grounding, modeled on interaction with an e-commerce website. We performed an empirical evaluation of autonomous agents trained using imitation and reinforcement learning,  and demonstrated promising results on sim-to-real transfer to real-world shopping websites.
Our qualitative and quantitative analysis of model and human trajectories (\S\ref{sec:breakdown}) identified several research challenges in \benchmark{} and provided insights for future model development by incorporating multidisciplinary techniques. 
For example, pre-training with multi-modal data~\cite{li2020oscar,wang2021simvlm}, web hypertext~\cite{Aghajanyan2021HTLMHP}, or web instruction-action mapping~\cite{pasupat2018elements} could help agents better understand and leverage rich semantics of webpage content, actions, and instructions. 
Ideas from query (re)formulation~\cite{komeili2021internet,zhuang2022bridging,nogueira2017task,wang2020deep} may help agents expand the range of search exploration, and improved action exploration~\cite{pathak2017curiosity,ecoffet2019go,tuyls2022multi} and memory~\cite{wayne2018unsupervised,fortunato2019generalization,lampinen2021towards} mechanisms could help agents make better decisions over the long horizon and large action space.
The modular design of \benchmark{} also allows for new web tasks and domains to be easily incorporated, which we hope will help shape future research into grounded language agents with stronger capabilities for real-world web interaction.

\section*{Acknowledgements}
\vspace{-5pt}
We thank Alexander Wettig, Ameet Deshpande, Austin Wang, Jens Tuyls, Jimmy Yang, Mengzhou Xia, Tianyu Gao, and Vishvak Murahari from the Princeton NLP Group for proofreading and providing comments.
This material is based upon work supported by the National Science Foundation under Grant No. 2107048. Any opinions, findings, and conclusions or recommendations expressed in this material are those of the author(s) and do not necessarily reflect the views of the National Science Foundation.

\bibliography{main}

\begin{thebibliography}{59}
\providecommand{\natexlab}[1]{#1}
\providecommand{\url}[1]{\texttt{#1}}
\expandafter\ifx\csname urlstyle\endcsname\relax
  \providecommand{\doi}[1]{doi: #1}\else
  \providecommand{\doi}{doi: \begingroup \urlstyle{rm}\Url}\fi

\bibitem[Adolphs et~al.(2021)Adolphs, Boerschinger, Buck, Huebscher, Ciaramita,
  Espeholt, Hofmann, and Kilcher]{adolphs2021boosting}
L.~Adolphs, B.~Boerschinger, C.~Buck, M.~C. Huebscher, M.~Ciaramita,
  L.~Espeholt, T.~Hofmann, and Y.~Kilcher.
\newblock {Boosting Search Engines with Interactive Agents}.
\newblock \emph{arXiv preprint arXiv:2109.00527}, 2021.

\bibitem[Aghajanyan et~al.(2021)Aghajanyan, Okhonko, Lewis, Joshi, Xu, Ghosh,
  and Zettlemoyer]{Aghajanyan2021HTLMHP}
A.~Aghajanyan, D.~Okhonko, M.~Lewis, M.~Joshi, H.~Xu, G.~Ghosh, and
  L.~Zettlemoyer.
\newblock Htlm: Hyper-text pre-training and prompting of language models.
\newblock \emph{ArXiv}, abs/2107.06955, 2021.

\bibitem[Andreas et~al.(2020)Andreas, Bufe, Burkett, Chen, Clausman, Crawford,
  Crim, DeLoach, Dorner, Eisner, et~al.]{andreas2020task}
J.~Andreas, J.~Bufe, D.~Burkett, C.~Chen, J.~Clausman, J.~Crawford, K.~Crim,
  J.~DeLoach, L.~Dorner, J.~Eisner, et~al.
\newblock Task-oriented dialogue as dataflow synthesis.
\newblock \emph{Transactions of the Association for Computational Linguistics},
  8:\penalty0 556--571, 2020.

\bibitem[Bender and Koller(2020)]{bender2020climbing}
E.~M. Bender and A.~Koller.
\newblock {Climbing towards NLU: On Meaning, Form, and Understanding in the Age
  of Data}.
\newblock In \emph{Proceedings of the 58th Annual Meeting of the Association
  for Computational Linguistics}, pages 5185--5198, 2020.

\bibitem[Brockman et~al.(2016)Brockman, Cheung, Pettersson, Schneider,
  Schulman, Tang, and Zaremba]{brockman2016openai}
G.~Brockman, V.~Cheung, L.~Pettersson, J.~Schneider, J.~Schulman, J.~Tang, and
  W.~Zaremba.
\newblock {OpenAI Gym}.
\newblock \emph{arXiv preprint arXiv:1606.01540}, 2016.

\bibitem[Brown et~al.(2020)Brown, Mann, Ryder, Subbiah, Kaplan, Dhariwal,
  Neelakantan, Shyam, Sastry, Askell, et~al.]{brown2020language}
T.~Brown, B.~Mann, N.~Ryder, M.~Subbiah, J.~D. Kaplan, P.~Dhariwal,
  A.~Neelakantan, P.~Shyam, G.~Sastry, A.~Askell, et~al.
\newblock Language models are few-shot learners.
\newblock \emph{Advances in neural information processing systems},
  33:\penalty0 1877--1901, 2020.

\bibitem[Budzianowski et~al.(2018)Budzianowski, Wen, Tseng, Casanueva, Ultes,
  Ramadan, and Ga{\v{s}}i{\'c}]{budzianowski2018multiwoz}
P.~Budzianowski, T.-H. Wen, B.-H. Tseng, I.~Casanueva, S.~Ultes, O.~Ramadan,
  and M.~Ga{\v{s}}i{\'c}.
\newblock Multiwoz--a large-scale multi-domain wizard-of-oz dataset for
  task-oriented dialogue modelling.
\newblock \emph{arXiv preprint arXiv:1810.00278}, 2018.

\bibitem[Burns et~al.(2022)Burns, Arsan, Agrawal, Kumar, Saenko, and
  Plummer]{burns2022interactive}
A.~Burns, D.~Arsan, S.~Agrawal, R.~Kumar, K.~Saenko, and B.~A. Plummer.
\newblock {Interactive Mobile App Navigation with Uncertain or Under-specified
  Natural Language Commands}.
\newblock \emph{arXiv preprint arXiv:2202.02312}, 2022.

\bibitem[Chung et~al.(2014)Chung, Çaglar G{\"u}lçehre, Cho, and
  Bengio]{Chung2014EmpiricalEO}
J.~Chung, Çaglar G{\"u}lçehre, K.~Cho, and Y.~Bengio.
\newblock Empirical evaluation of gated recurrent neural networks on sequence
  modeling.
\newblock \emph{ArXiv}, abs/1412.3555, 2014.

\bibitem[Devlin et~al.(2019{\natexlab{a}})Devlin, Chang, Lee, and
  Toutanova]{Devlin2019BERTPO}
J.~Devlin, M.-W. Chang, K.~Lee, and K.~Toutanova.
\newblock {BERT: Pre-training of Deep Bidirectional Transformers for Language
  Understanding}.
\newblock \emph{ArXiv}, abs/1810.04805, 2019{\natexlab{a}}.

\bibitem[Devlin et~al.(2019{\natexlab{b}})Devlin, Chang, Lee, and
  Toutanova]{devlin2019bert}
J.~Devlin, M.-W. Chang, K.~Lee, and K.~Toutanova.
\newblock Bert: Pre-training of deep bidirectional transformers for language
  understanding.
\newblock In \emph{NAACL-HLT (1)}, 2019{\natexlab{b}}.

\bibitem[Ecoffet et~al.(2019)Ecoffet, Huizinga, Lehman, Stanley, and
  Clune]{ecoffet2019go}
A.~Ecoffet, J.~Huizinga, J.~Lehman, K.~O. Stanley, and J.~Clune.
\newblock Go-explore: a new approach for hard-exploration problems.
\newblock \emph{arXiv preprint arXiv:1901.10995}, 2019.

\bibitem[Fortunato et~al.(2019)Fortunato, Tan, Faulkner, Hansen,
  Puigdom{\`e}nech~Badia, Buttimore, Deck, Leibo, and
  Blundell]{fortunato2019generalization}
M.~Fortunato, M.~Tan, R.~Faulkner, S.~Hansen, A.~Puigdom{\`e}nech~Badia,
  G.~Buttimore, C.~Deck, J.~Z. Leibo, and C.~Blundell.
\newblock Generalization of reinforcement learners with working and episodic
  memory.
\newblock \emph{Advances in neural information processing systems}, 32, 2019.

\bibitem[Guo et~al.(2020)Guo, Yu, Gao, Gan, Campbell, and
  Chang]{guo2020interactive}
X.~Guo, M.~Yu, Y.~Gao, C.~Gan, M.~Campbell, and S.~Chang.
\newblock Interactive fiction game playing as multi-paragraph reading
  comprehension with reinforcement learning.
\newblock \emph{arXiv preprint arXiv:2010.02386}, 2020.

\bibitem[Gur et~al.(2018)Gur, Rueckert, Faust, and
  Hakkani-Tur]{gur2018learning}
I.~Gur, U.~Rueckert, A.~Faust, and D.~Hakkani-Tur.
\newblock {Learning to Navigate the Web}.
\newblock \emph{arXiv preprint arXiv:1812.09195}, 2018.

\bibitem[Gur et~al.(2021)Gur, Jaques, Malta, Tiwari, Lee, and
  Faust]{gur2021adversarial}
I.~Gur, N.~Jaques, K.~Malta, M.~Tiwari, H.~Lee, and A.~Faust.
\newblock {Adversarial Environment Generation for Learning to Navigate the
  Web}.
\newblock \emph{arXiv preprint arXiv:2103.01991}, 2021.

\bibitem[Hausknecht et~al.(2020)Hausknecht, Ammanabrolu, C{\^o}t{\'e}, and
  Yuan]{hausknecht2020interactive}
M.~Hausknecht, P.~Ammanabrolu, M.-A. C{\^o}t{\'e}, and X.~Yuan.
\newblock Interactive fiction games: A colossal adventure.
\newblock In \emph{Proceedings of the AAAI Conference on Artificial
  Intelligence}, volume~34, pages 7903--7910, 2020.

\bibitem[He et~al.(2016)He, Zhang, Ren, and Sun]{He2016DeepRL}
K.~He, X.~Zhang, S.~Ren, and J.~Sun.
\newblock {Deep Residual Learning for Image Recognition}.
\newblock \emph{2016 IEEE Conference on Computer Vision and Pattern Recognition
  (CVPR)}, pages 770--778, 2016.

\bibitem[Hotti et~al.(2021)Hotti, Risuleo, Magureanu, Moradi, and
  Lagergren]{hotti2021klarna}
A.~Hotti, R.~S. Risuleo, S.~Magureanu, A.~Moradi, and J.~Lagergren.
\newblock {The Klarna Product Page Dataset: A RealisticBenchmark for Web
  Representation Learning}.
\newblock \emph{arXiv preprint arXiv:2111.02168}, 2021.

\bibitem[Humphreys et~al.(2022)Humphreys, Raposo, Pohlen, Thornton, Chhaparia,
  Muldal, Abramson, Georgiev, Goldin, Santoro, et~al.]{humphreys2022data}
P.~C. Humphreys, D.~Raposo, T.~Pohlen, G.~Thornton, R.~Chhaparia, A.~Muldal,
  J.~Abramson, P.~Georgiev, A.~Goldin, A.~Santoro, et~al.
\newblock {A data-driven approach for learning to control computers}.
\newblock \emph{arXiv preprint arXiv:2202.08137}, 2022.

\bibitem[Jia et~al.(2019)Jia, Kiros, and Ba]{jia2019dom}
S.~Jia, J.~Kiros, and J.~Ba.
\newblock {Dom-q-net: Grounded RL on Structured Language}.
\newblock \emph{arXiv preprint arXiv:1902.07257}, 2019.

\bibitem[Komeili et~al.(2021)Komeili, Shuster, and Weston]{komeili2021internet}
M.~Komeili, K.~Shuster, and J.~Weston.
\newblock Internet-augmented dialogue generation.
\newblock \emph{arXiv preprint arXiv:2107.07566}, 2021.

\bibitem[Lampinen et~al.(2021)Lampinen, Chan, Banino, and
  Hill]{lampinen2021towards}
A.~Lampinen, S.~Chan, A.~Banino, and F.~Hill.
\newblock Towards mental time travel: a hierarchical memory for reinforcement
  learning agents.
\newblock \emph{Advances in Neural Information Processing Systems},
  34:\penalty0 28182--28195, 2021.

\bibitem[Lazaridou et~al.(2020)Lazaridou, Potapenko, and
  Tieleman]{Lazaridou2020MultiagentCM}
A.~Lazaridou, A.~Potapenko, and O.~Tieleman.
\newblock {Multi-agent Communication meets Natural Language: Synergies between
  Functional and Structural Language Learning}.
\newblock In \emph{ACL}, 2020.

\bibitem[Lazaridou et~al.(2022)Lazaridou, Gribovskaya, Stokowiec, and
  Grigorev]{Lazaridou2022InternetaugmentedLM}
A.~Lazaridou, E.~Gribovskaya, W.~Stokowiec, and N.~Grigorev.
\newblock Internet-augmented language models through few-shot prompting for
  open-domain question answering.
\newblock \emph{ArXiv}, abs/2203.05115, 2022.

\bibitem[Lewis et~al.(2019)Lewis, Liu, Goyal, Ghazvininejad, Mohamed, Levy,
  Stoyanov, and Zettlemoyer]{lewis2019bart}
M.~Lewis, Y.~Liu, N.~Goyal, M.~Ghazvininejad, A.~Mohamed, O.~Levy, V.~Stoyanov,
  and L.~Zettlemoyer.
\newblock {BART: Denoising Sequence-to-Sequence Pre-training for Natural
  Language Generation, Translation, and Comprehension}.
\newblock \emph{arXiv preprint arXiv:1910.13461}, 2019.

\bibitem[Li et~al.(2020)Li, Yin, Li, Zhang, Hu, Zhang, Wang, Hu, Dong, Wei,
  et~al.]{li2020oscar}
X.~Li, X.~Yin, C.~Li, P.~Zhang, X.~Hu, L.~Zhang, L.~Wang, H.~Hu, L.~Dong,
  F.~Wei, et~al.
\newblock Oscar: Object-semantics aligned pre-training for vision-language
  tasks.
\newblock In \emph{European Conference on Computer Vision}, pages 121--137.
  Springer, 2020.

\bibitem[Lin et~al.(2021)Lin, Ma, Lin, Yang, Pradeep, and
  Nogueira]{lin2021pyserini}
J.~Lin, X.~Ma, S.-C. Lin, J.-H. Yang, R.~Pradeep, and R.~Nogueira.
\newblock {Pyserini: An Easy-to-Use Python Toolkit to Support Replicable IR
  Research with Sparse and Dense Representationss}.
\newblock \emph{arXiv preprint arXiv:2102.10073}, 2021.

\bibitem[Liu et~al.(2018)Liu, Guu, Pasupat, Shi, and
  Liang]{liu2018reinforcement}
E.~Z. Liu, K.~Guu, P.~Pasupat, T.~Shi, and P.~Liang.
\newblock {Reinforcement Learning on Web Interfaces using Workflow-Guided
  Exploration}.
\newblock \emph{arXiv preprint arXiv:1802.08802}, 2018.

\bibitem[Luketina et~al.(2019)Luketina, Nardelli, Farquhar, Foerster, Andreas,
  Grefenstette, Whiteson, and Rockt{\"a}schel]{luketina2019survey}
J.~Luketina, N.~Nardelli, G.~Farquhar, J.~N. Foerster, J.~Andreas,
  E.~Grefenstette, S.~Whiteson, and T.~Rockt{\"a}schel.
\newblock A survey of reinforcement learning informed by natural language.
\newblock In \emph{IJCAI}, 2019.

\bibitem[Mazumder and Riva(2020)]{mazumder2020flin}
S.~Mazumder and O.~Riva.
\newblock {FLIN: A Flexible Natural Language Interface for Web Navigation}.
\newblock \emph{arXiv preprint arXiv:2010.12844}, 2020.

\bibitem[Mnih et~al.(2016)Mnih, Badia, Mirza, Graves, Lillicrap, Harley,
  Silver, and Kavukcuoglu]{mnih2016asynchronous}
V.~Mnih, A.~P. Badia, M.~Mirza, A.~Graves, T.~Lillicrap, T.~Harley, D.~Silver,
  and K.~Kavukcuoglu.
\newblock Asynchronous methods for deep reinforcement learning.
\newblock In \emph{International conference on machine learning}, pages
  1928--1937. PMLR, 2016.

\bibitem[Nakano et~al.(2021)Nakano, Hilton, Balaji, Wu, Ouyang, Kim, Hesse,
  Jain, Kosaraju, Saunders, et~al.]{nakano2021webgpt}
R.~Nakano, J.~Hilton, S.~Balaji, J.~Wu, L.~Ouyang, C.~Kim, C.~Hesse, S.~Jain,
  V.~Kosaraju, W.~Saunders, et~al.
\newblock {WebGPT: Browser-Assisted Question-Answering with Human Feedback}.
\newblock \emph{arXiv preprint arXiv:2112.09332}, 2021.

\bibitem[Narasimhan et~al.(2016)Narasimhan, Yala, and
  Barzilay]{narasimhan2016improving}
K.~Narasimhan, A.~Yala, and R.~Barzilay.
\newblock {Improving Information Extraction by Acquiring External Evidence with
  Reinforcement Learning}.
\newblock In \emph{Proceedings of the 2016 Conference on Empirical Methods in
  Natural Language Processing}, pages 2355--2365, 2016.

\bibitem[Ni(2015)]{scraperapi}
D.~Ni.
\newblock {ScraperAPI}, 2015.
\newblock URL \url{https://www.scraperapi.com/}.

\bibitem[Nogueira and Cho(2016)]{nogueira2016end}
R.~Nogueira and K.~Cho.
\newblock {End-to-End Goal-Driven Web Navigation}.
\newblock \emph{{Advances in Neural Information Processing Systems}}, 29, 2016.

\bibitem[Nogueira and Cho(2017)]{nogueira2017task}
R.~Nogueira and K.~Cho.
\newblock {Task-Oriented Query Reformulation with Reinforcement Learning}.
\newblock In \emph{Proceedings of the 2017 Conference on Empirical Methods in
  Natural Language Processing}, pages 574--583, 2017.

\bibitem[Pasupat et~al.(2018{\natexlab{a}})Pasupat, Jiang, Liu, Guu, and
  Liang]{pasupat2018elements}
P.~Pasupat, T.-S. Jiang, E.~Z. Liu, K.~Guu, and P.~Liang.
\newblock Mapping natural language commands to web elements.
\newblock In \emph{Empirical Methods in Natural Language Processing (EMNLP)},
  2018{\natexlab{a}}.

\bibitem[Pasupat et~al.(2018{\natexlab{b}})Pasupat, Jiang, Liu, Guu, and
  Liang]{pasupat2018mapping}
P.~Pasupat, T.-S. Jiang, E.~Z. Liu, K.~Guu, and P.~Liang.
\newblock Mapping natural language commands to web elements.
\newblock In \emph{EMNLP}, 2018{\natexlab{b}}.

\bibitem[Pathak et~al.(2017)Pathak, Agrawal, Efros, and
  Darrell]{pathak2017curiosity}
D.~Pathak, P.~Agrawal, A.~A. Efros, and T.~Darrell.
\newblock Curiosity-driven exploration by self-supervised prediction.
\newblock In \emph{International conference on machine learning}, pages
  2778--2787. PMLR, 2017.

\bibitem[Ronacher(2010)]{flask}
A.~Ronacher.
\newblock {Flask API}, 2010.
\newblock URL \url{https://flask.palletsprojects.com/en/2.1.x/}.

\bibitem[Seo et~al.(2016)Seo, Kembhavi, Farhadi, and
  Hajishirzi]{seo2016bidirectional}
M.~Seo, A.~Kembhavi, A.~Farhadi, and H.~Hajishirzi.
\newblock Bidirectional attention flow for machine comprehension.
\newblock \emph{arXiv preprint arXiv:1611.01603}, 2016.

\bibitem[Shi et~al.(2017)Shi, Karpathy, Fan, Hernandez, and
  Liang]{shi2017world}
T.~Shi, A.~Karpathy, L.~Fan, J.~Hernandez, and P.~Liang.
\newblock {World of Bits: An Open-Domain platform for web-based agents}.
\newblock In \emph{International Conference on Machine Learning}, pages
  3135--3144. PMLR, 2017.

\bibitem[Shridhar et~al.(2020)Shridhar, Thomason, Gordon, Bisk, Han, Mottaghi,
  Zettlemoyer, and Fox]{shridhar2020alfred}
M.~Shridhar, J.~Thomason, D.~Gordon, Y.~Bisk, W.~Han, R.~Mottaghi,
  L.~Zettlemoyer, and D.~Fox.
\newblock Alfred: A benchmark for interpreting grounded instructions for
  everyday tasks.
\newblock In \emph{Proceedings of the IEEE/CVF conference on computer vision
  and pattern recognition}, pages 10740--10749, 2020.

\bibitem[Shuster et~al.(2022)Shuster, Komeili, Adolphs, Roller, Szlam, and
  Weston]{Shuster2022LanguageMT}
K.~Shuster, M.~Komeili, L.~Adolphs, S.~Roller, A.~D. Szlam, and J.~Weston.
\newblock Language models that seek for knowledge: Modular search \& generation
  for dialogue and prompt completion.
\newblock \emph{ArXiv}, abs/2203.13224, 2022.

\bibitem[Su et~al.(2017)Su, Awadallah, Khabsa, Pantel, Gamon, and
  Encarnacion]{su2017building}
Y.~Su, A.~H. Awadallah, M.~Khabsa, P.~Pantel, M.~Gamon, and M.~Encarnacion.
\newblock {Building Natural Language Interfaces to Web APIs}.
\newblock In \emph{Proceedings of the 2017 ACM on Conference on Information and
  Knowledge Management}, pages 177--186, 2017.

\bibitem[Su et~al.(2018)Su, Hassan~Awadallah, Wang, and White]{su2018natural}
Y.~Su, A.~Hassan~Awadallah, M.~Wang, and R.~W. White.
\newblock {Natural Language Interfaces with Fine-Grained User Interaction: A
  Case Study on Web APIs}.
\newblock In \emph{The 41st International ACM SIGIR Conference on Research \&
  Development in Information Retrieval}, pages 855--864, 2018.

\bibitem[Toyama et~al.(2021)Toyama, Hamel, Gergely, Comanici, Glaese, Ahmed,
  Jackson, Mourad, and Precup]{toyama2021androidenv}
D.~Toyama, P.~Hamel, A.~Gergely, G.~Comanici, A.~Glaese, Z.~Ahmed, T.~Jackson,
  S.~Mourad, and D.~Precup.
\newblock {AndroidEnv: A Reinforcement Learning Platform for Android}.
\newblock \emph{arXiv preprint arXiv:2105.13231}, 2021.

\bibitem[Tuyls et~al.(2022)Tuyls, Yao, Kakade, and Narasimhan]{tuyls2022multi}
J.~Tuyls, S.~Yao, S.~Kakade, and K.~Narasimhan.
\newblock Multi-stage episodic control for strategic exploration in text games.
\newblock \emph{arXiv preprint arXiv:2201.01251}, 2022.

\bibitem[Uc-Cetina et~al.(2021)Uc-Cetina, Navarro-Guerrero, Martin-Gonzalez,
  Weber, and Wermter]{uc2021survey}
V.~Uc-Cetina, N.~Navarro-Guerrero, A.~Martin-Gonzalez, C.~Weber, and
  S.~Wermter.
\newblock Survey on reinforcement learning for language processing.
\newblock \emph{arXiv preprint arXiv:2104.05565}, 2021.

\bibitem[Wang et~al.(2020)Wang, Macdonald, and Ounis]{wang2020deep}
X.~Wang, C.~Macdonald, and I.~Ounis.
\newblock Deep reinforced query reformulation for information retrieval.
\newblock \emph{arXiv preprint arXiv:2007.07987}, 2020.

\bibitem[Wang et~al.(2021)Wang, Yu, Yu, Dai, Tsvetkov, and Cao]{wang2021simvlm}
Z.~Wang, J.~Yu, A.~W. Yu, Z.~Dai, Y.~Tsvetkov, and Y.~Cao.
\newblock Simvlm: Simple visual language model pretraining with weak
  supervision.
\newblock \emph{arXiv preprint arXiv:2108.10904}, 2021.

\bibitem[Wayne et~al.(2018)Wayne, Hung, Amos, Mirza, Ahuja, Grabska-Barwinska,
  Rae, Mirowski, Leibo, Santoro, et~al.]{wayne2018unsupervised}
G.~Wayne, C.-C. Hung, D.~Amos, M.~Mirza, A.~Ahuja, A.~Grabska-Barwinska,
  J.~Rae, P.~Mirowski, J.~Z. Leibo, A.~Santoro, et~al.
\newblock Unsupervised predictive memory in a goal-directed agent.
\newblock \emph{arXiv preprint arXiv:1803.10760}, 2018.

\bibitem[Williams et~al.(2019)Williams, Hashemi, and
  Zitouni]{williams2019automatic}
K.~Williams, S.~H. Hashemi, and I.~Zitouni.
\newblock {Automatic Task Completion Flows from Web APIs}.
\newblock In \emph{Proceedings of the 42nd International ACM SIGIR Conference
  on Research and Development in Information Retrieval}, pages 1009--1012,
  2019.

\bibitem[Yao et~al.(2020)Yao, Rao, Hausknecht, and Narasimhan]{Yao2020KeepCA}
S.~Yao, R.~Rao, M.~J. Hausknecht, and K.~Narasimhan.
\newblock {Keep CALM and Explore: Language Models for Action Generation in
  Text-based Games}.
\newblock \emph{ArXiv}, abs/2010.02903, 2020.

\bibitem[Yao et~al.(2021)Yao, Narasimhan, and Hausknecht]{yao2021reading}
S.~Yao, K.~Narasimhan, and M.~Hausknecht.
\newblock Reading and acting while blindfolded: The need for semantics in text
  game agents.
\newblock \emph{arXiv preprint arXiv:2103.13552}, 2021.

\bibitem[Yuan et~al.(2020)Yuan, Fu, C{\^o}t{\'e}, Tay, Pal, and
  Trischler]{Yuan2020InteractiveMC}
X.~Yuan, J.~Fu, M.-A. C{\^o}t{\'e}, Y.~Tay, C.~J. Pal, and A.~Trischler.
\newblock Interactive machine comprehension with information seeking agents.
\newblock In \emph{ACL}, 2020.

\bibitem[Zhong et~al.(2021)Zhong, Hanjie, Wang, Narasimhan, and
  Zettlemoyer]{zhong2021silg}
V.~Zhong, A.~W. Hanjie, S.~Wang, K.~Narasimhan, and L.~Zettlemoyer.
\newblock Silg: The multi-domain symbolic interactive language grounding
  benchmark.
\newblock \emph{Advances in Neural Information Processing Systems},
  34:\penalty0 21505--21519, 2021.

\bibitem[Zhuang et~al.(2022)Zhuang, Ren, Shou, Pei, Gong, Zuccon, and
  Jiang]{zhuang2022bridging}
S.~Zhuang, H.~Ren, L.~Shou, J.~Pei, M.~Gong, G.~Zuccon, and D.~Jiang.
\newblock Bridging the gap between indexing and retrieval for differentiable
  search index with query generation.
\newblock \emph{arXiv preprint arXiv:2206.10128}, 2022.

\end{thebibliography}
\bibliographystyle{abbrvnat}

\section*{Checklist}

\begin{enumerate}

\item For all authors...
\begin{enumerate}
  \item Do the main claims made in the abstract and introduction accurately reflect the paper's contributions and scope?
    \answerYes{}
  \item Did you describe the limitations of your work?
    \answerYes{} See Section 6 Discussion and Appendix.
  \item Did you discuss any potential negative societal impacts of your work?
    \answerYes{} See Section 6 Discussion and Appendix.
  \item Have you read the ethics review guidelines and ensured that your paper conforms to them?
    \answerYes{}
\end{enumerate}

\item If you are including theoretical results...
\begin{enumerate}
  \item Did you state the full set of assumptions of all theoretical results?
    \answerNA{}
        \item Did you include complete proofs of all theoretical results?
    \answerNA{}
\end{enumerate}

\item If you ran experiments...
\begin{enumerate}
  \item Did you include the code, data, and instructions needed to reproduce the main experimental results (either in the supplemental material or as a URL)?
    \answerYes{} See supplementary materials.
  \item Did you specify all the training details (e.g., data splits, hyperparameters, how they were chosen)?
    \answerYes{} Data splits are described in the Section 5 first paragraph. Hyperparameters and training details are in the Appendix.
        \item Did you report error bars (e.g., with respect to the random seed after running experiments multiple times)?
    \answerYes{} Figure 3 includes error bars, Table 2 includes min/max statistics along with averages.
        \item Did you include the total amount of compute and the type of resources used (e.g., type of GPUs, internal cluster, or cloud provider)?
    \answerYes{} In appendix training details.
\end{enumerate}

\item If you are using existing assets (e.g., code, data, models) or curating/releasing new assets...
\begin{enumerate}
  \item If your work uses existing assets, did you cite the creators?
    \answerYes{} Citations include ScraperAPI, Flask, OpenAI Gym, BERT, BART, A2C.
  \item Did you mention the license of the assets?
    \answerYes{} Discussed in appendix.
  \item Did you include any new assets either in the supplemental material or as a URL?
    \answerYes{} In the supplementary materials.
  \item Did you discuss whether and how consent was obtained from people whose data you're using/curating?
    \answerYes{} Discussed in appendix, we only scrape publicly available data from the Internet.
  \item Did you discuss whether the data you are using/curating contains personally identifiable information or offensive content?
    \answerYes{} Discussed in Appendix.
\end{enumerate}

\item If you used crowdsourcing or conducted research with human subjects...
\begin{enumerate}
  \item Did you include the full text of instructions given to participants and screenshots, if applicable?
    \answerYes{} In appendix.
  \item Did you describe any potential participant risks, with links to Institutional Review Board (IRB) approvals, if applicable?
    \answerYes{} Discussed in Appendix.
  \item Did you include the estimated hourly wage paid to participants and the total amount spent on participant compensation?
    \answerYes{} Discussed in Appendix.
\end{enumerate}

\end{enumerate}

\newpage
\appendix

\section{Environment Details}\label{appx:env}

\subsection{Product Scraping}
\label{appx:scrape}
We use ScraperAPI~\cite{scraperapi} to extract publicly available product information from \url{amazon.com}. We use five categories (beauty, food, fashion, furniture, electronics) and 313 associated sub-category names appeared in \url{amazon.com} (e.g.\,``Women's Loafers \& Slip-Ons'' in fashion, ``Pendants and Chandeliers'' in furniture) to scrape $1,181,436$ products. We filter products with duplicate titles or product IDs, but do not perform extra filtering in order to avoid selection bias. Specifically, as \url{amazon.com} has its own content screening process, we did not find any personally identifiable information or offensive content during random sampling checks.

\begin{table}[ht]
\centering
\begin{tabular}{ccccc}
\toprule
    Products & Unique Attributes & Avg Attributes & Unique Options  & Avg Options\\
\midrule
    1,181,436 & 670 & 3.1 & 842,849 & 0.67\\
\bottomrule
\end{tabular}
\vspace{5pt}
\caption{Product item statistics.}
\label{table:product_stats}
\end{table}

\subsection{Product Attribute Mining}
\label{appx:attribute}

We use \texttt{TfidfVectorizer} from \texttt{scikit-learn} to extract probable bi-grams as attributes from product title and descriptions for further annotation. We manually inspect these attributes to keep only the \textit{specific} and \textit{human-readable} ones and filter out the rest. An attribute should be suitable in at least one of the following use: 1) \texttt{IsGoodFor}, 2) \texttt{HasA} (contains), 3) \texttt{WhichIs}, and 4) \texttt{IsA}.
For example, attributes such as ``oz ml'' and ``men women'' will be filtered out since it's unparsable. On the other hand, ``hair color'' will also be filtered since it is not specific enough to fit in the above $4$ categories. Attributes such as ``dry skin'' can fit the \texttt{IsGoodFor} in the context of a make-up product being good for dry skin.

\subsection{Search Engine}
\label{appx:search_engine}

{\color{black}{Each time the agent performs a search, the top 50 items are retrieved and displayed across five search result pages, where each page contains 10 items and the agent can use actions \texttt{choose}[Prev/Next page] to navigate across result pages. Figure~\ref{fig:search} shows that when searching directly with the instruction text, the corresponding item appears in the first search page (rank 1-10) nearly 1/3 of the time, but it cannot be found in any search pages (rank 50+) more than half of the time. This indicates that while the search engine can decently retrieve items based on lexical matching, directly searching the instruction is not enough for solving the task, and good query (re)formulation based on the instruction is important.}}

\subsection{Instruction Collection}
\label{appx:instruction}

\begin{figure}[t!]
    \centering
    \includegraphics[width=1.0\textwidth]{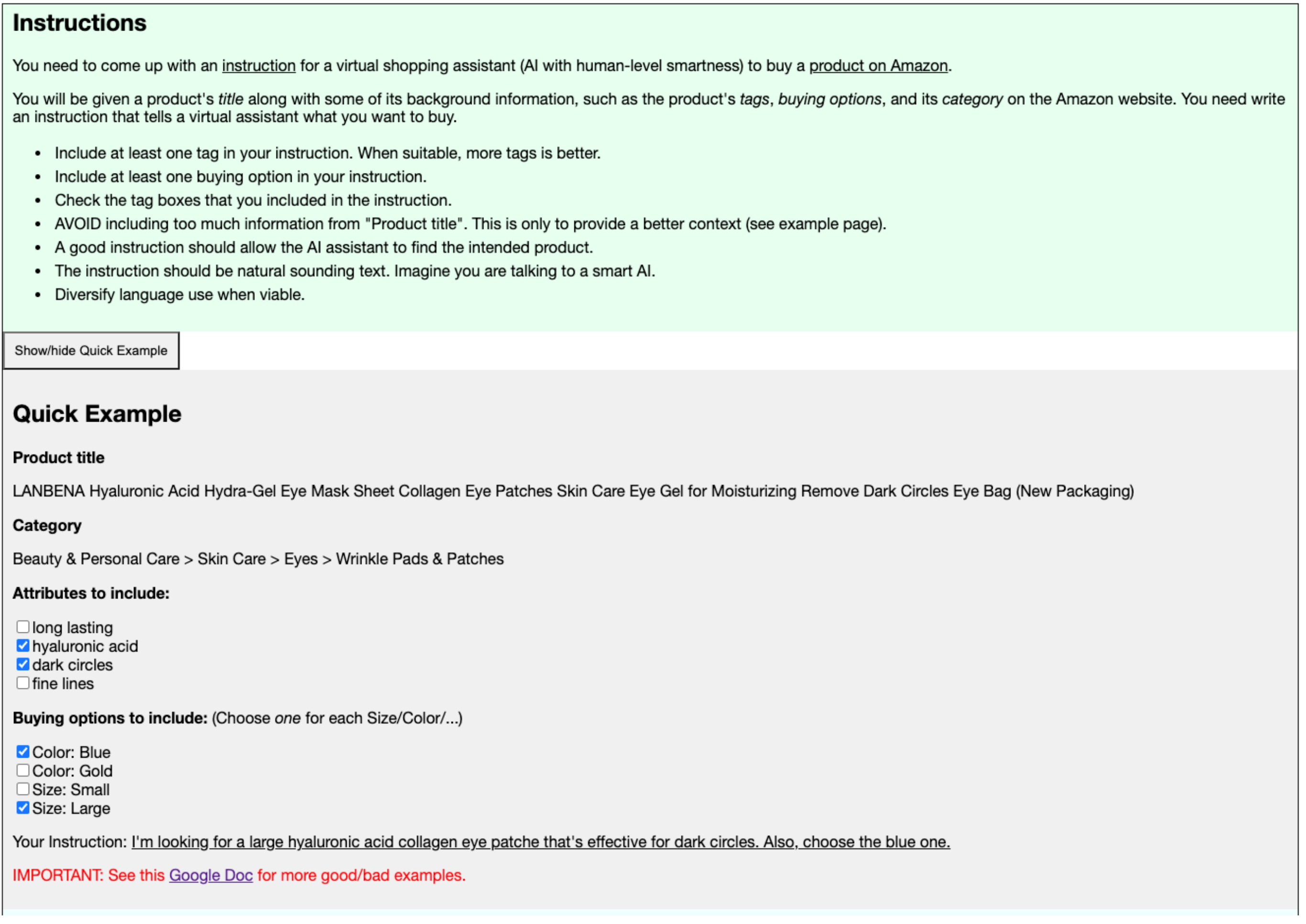}
    \caption{The Amazon Mechanical Turk interface for the instruction writing task. The green box shows the general instruction for the task and the grey box shows an concrete example.}
    \label{fig:mturk_inst_example} 
\end{figure}

\begin{figure}[t!]
    \centering
    \includegraphics[width=1.0\textwidth]{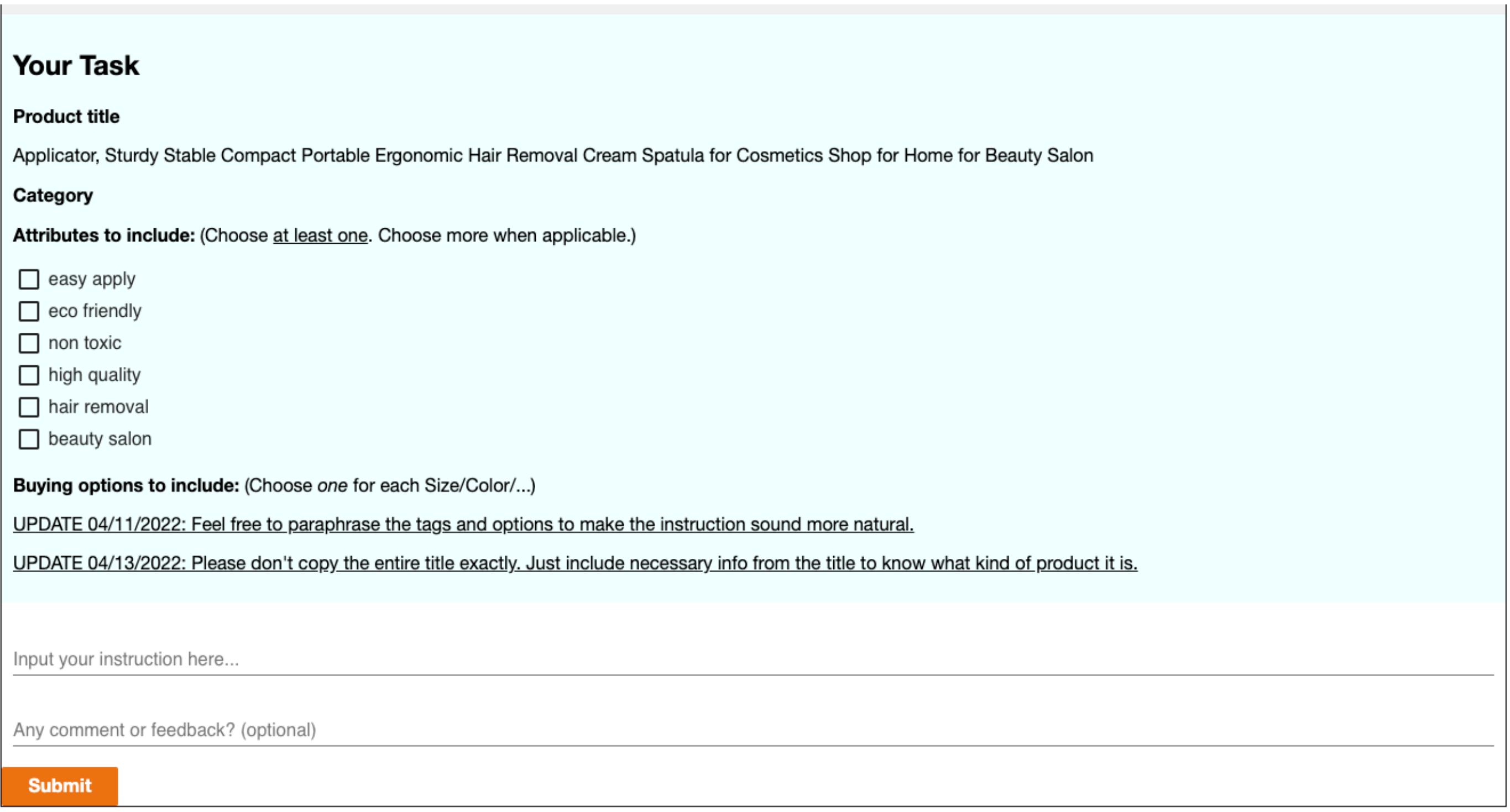}
    \caption{The Amazon Mechanical Turk interface for the instruction writing task. The blue box shows the actual annotation interface. The worker is required to check the boxes and write the instructions in the text field before submission.}
    \label{fig:mturk_inst_task} 
\end{figure}

We collect human written instructions by providing the workers a product including the title, product category, and its set of attributes and options (Figure~\ref{fig:mturk_inst_example}, \ref{fig:mturk_inst_task}).
We conduct qualification task by having each participating workers to work on $2 - 5$ examples. We inspect and assign qualification to $213$ workers to perform the instruction writing task. We pay for each example $0.15$ dollars. We do not anticipate any potential participant risk.

\subsection{Reward Calculation}
\label{rewardCalc}
The type reward $r_\text{type}$ consists of $3$ elements: 1) course-grain product category match ($c = 1$ if matched), 2) fine-grain category match ($f = 1$ if matched), and 3) product title match. Course-grain product category refers to the $5$ categories described in \S\ref{sec:task}. Fine-grain category is the chain of categories that the product is under on the Amazon website. For example, and eye mask sheet would be under the \textit{Beauty \& Personal Care > Skin Care > Eyes > Wrinkle Pads \& Patches} fine-grain category. The product title refers to $\ptext$ described in \S\ref{sec:env}. 

\begin{equation}
    r_{\text{type}}= 
\begin{cases}
    0,& \text{if }\texttt{TextMatch}(\ptext, \ptext^*) = 0\\
    0.1,& \text{if }\texttt{TextMatch}(\ptext, \ptext^*) < 0.1\\
    0.5,& \text{if }\texttt{TextMatch}(\ptext, \ptext^*) > 0.2\text{ and }c=1\text{ and }f=1,\\
    1,              & \text{otherwise}
\end{cases}
\end{equation}
Here, \texttt{TextMatch}($\ptext$, $\ptext^*$) is a simple string match between the selected product title text and the goal product title text. We use only the words tagged with PNOUN, NOUN, and PROPN tags parsed by the SpaCy parser in the title text.

\subsection{Human Trajectory Collection}
\label{appx:traj}
We use the \html{} environment in Figure~\ref{fig:teaser} to collect human trajectories. We select a pool of $13$ workers using qualification tasks where each workers complete $5$ examples. The workers that achieve an average reward more than $0.75$ are qualified. The task instruction is shown at the end of Appendix. We observe a pronounced performance gap between the very high performing workers and average workers. We use the top $50\%$ of these qualified workers as experts ($7$ workers in total). We pay for each completed trajectory $0.7$ dollars. {\color{black} In human evaluation, $8$ out of the $13$ workers participated and $5$ among them are in the aformentioned expert pool. The $8$ participants achieve an overall score of $75.5$ and a success rate of $50.0\%$
We observe non-negligible variance even within the experts---the best performer achieves a score of $87.4$ and success rate of $69.5\%$, while the lowest performing worker achieves a score of $45.8$ and success rate of $10\%$. The best performing worker also shows better consistency---drawing at a standard deviation of $2.3$ in score, contrasting the lowest performing counterpart at $3.1$. We provide examples of common human failure cases such as not matching the option/attribute due to impatience (Table \ref{table:bad_turker_traj_example}), cautioning some caveats of the task with human workers.}

\begin{table}[t]
\centering
\footnotesize

\begin{minipage}{.49\linewidth}
\begin{tabularx}{\textwidth}{X}
\toprule

{\centering\textbf{Instruction 1}}:
{I would like a \textcolor{mellowred}{stained glass} wall lamp with a \textcolor{mellowblue}{bronze finish}, and price lower than 190 dollars.}
\\ \midrule 

\textbf{Human Actions } ($r$~=~0.33, length~=~4)\\
 \textcolor{black}{search[\textcolor{mellowred}{stained glass} wall lamp]} 
click[item-QCLU Tiffany Style Lamp Sunflower...] 
click[wall lamp 3 - 12 inch]
click[buy] \\
\bottomrule
\end{tabularx}

\end{minipage}
\begin{minipage}{.49\linewidth}

\begin{tabularx}{\textwidth}{X}

\toprule
\textbf{Instruction 2}\\
{I would like a \textcolor{mellowblue}{lead free} \textcolor{mellowred}{bracelet} \textcolor{mellowred}{birthday cake} jar candle, and price lower than 50.00 dollars.
}
\\ \midrule 
\textbf{Human Actions} ($r$~=~0.03, len~=~4) \\
{search[\textcolor{mellowblue}{lead free} \textcolor{mellowred}{bracelet} \textcolor{mellowred}{birthday cake} jar candle]}
click[item-Happy Birthday Candle...]
click[8 ounce round tin]
click[buy] \\
\bottomrule
\end{tabularx}
\end{minipage}
\vspace{5pt}
\caption{\textcolor{black}{Two examples of failed human trajectories. A common pattern is impatience: after one search (even with correct attributes like the right example) the less performant worker commits to the first selected item. Often, the item does not contain the desired options even though the item's title text seem relevant. An expert worker will recognize the need to select the correct options and go back to refine the searches, while less performant workers simply commit to the current selected item.}}
\label{table:bad_turker_traj_example}
\end{table}

\subsection{Reward Verification}
\label{sec:reward_verification}
We randomly select $100$ samples each from the pools of trajectories generated by average and expert MTurk workers. Each trajectory is then manually re-scored against a human criteria; the purpose of this is to determine how representative the reward function is of a human's judgment towards whether the chosen product satisfies the given instructions. The human score calculation procedure exactly follows the formula laid out in Section \ref{rewardCalc} -- the attribute, option, price, and type scores are individually determined, then aggregated to calculate the overall score -- except for one main modification. Instead of the exact matching approach, points are awarded if (1) the picked product's attributes, options, or type are lexically similar or synonymous with the goal's product information and (2) the desired value is not found verbatim anywhere in the picked product's descriptions. For instance, if the value \textit{lightweight} is specified as a desired attribute for an instruction, but the value \textit{easy carry} is found instead in the picked product's description, then the attribute score for the picked product is increased to reflect that the \textit{lightweight} value was found. On the other hand, if \textit{cyan} is desired as an option for a goal product, but the user picks \textit{blue} even though \textit{cyan} is available as a choice, then no points are awarded. To ensure the score is calculated without bias, the original rewards for each trajectory were not compared with the human evaluation scores until the human evaluation scoring was completed.

{\color{black} For the average trajectories, the automatic task score was $74.9$ and our manual score was $76.3$ with a Pearson correlation of $0.856$. For expert trajectories, the respective scores were $81.5$ and $89.9$ with a Pearson correlation of $0.773$. Therefore, the automatic reward seems to provide a reasonably close lower bound to the actual task performance. We find that for average workers, $87.0\%$ of automatic scores are within a $10\%$ of the manual score, with the main source of error being synonyms or lexically similar words that don't get matched correctly in the automatic reward function.}

\begin{table}[ht]
\centering
\begin{tabular}{llccccc}
\toprule
    MTurk Type & Reward Function & Price & Type & Attribute & Result & Overall \\
\midrule
    Average & \benchmark{} & 95.0 & 92.9 & 71.7 & 50.5 & 74.9\\
            & Human        & 95.0 & 93.8 & 75.3 & 57.0 & 76.3\\
    Expert  & \benchmark{} & 100.0 & 100.0 & 78.1 & 56.1 & 81.5\\
            & Human        & 100.0 & 100.0 & 88.2 & 66.8 & 89.9\\
\bottomrule
\end{tabular}
\vspace{5pt}
\caption{Reward Verification Statistics}
\label{table:verify_stats}
\end{table}

Table \ref{table:verify_stats} reflects our observation that our reward function is similar to a human's score, with a consistent tendency to over-penalize the picked product. For every trajectory's product, the human score across all categories (e.g.\,attributes, options) is always greater than or equal to the original score. This under-scoring is a result of our reward function's exact matching criterion. In future work, we hope to improve our matching functionality such that, within the context of a single product with respect to the goal instructions, it can identify synonyms and decide whether to award additional points.

\section{Model Details}\label{appx:model}

\subsection{Cross Attention Layer}
Our cross attention layer follows \citet{seo2016bidirectional}. Denote the $i$-th contextualized token embedding from the observation and action to be $\mathbf{o}_i$ and $\mathbf{a}_i$ respectively. The attention between $\mathbf{o}_i$ and $\mathbf{a}_j$ is defined as 
\begin{equation}
    \mathbf{\alpha}_{ij} = \mathbf{w}_1 \cdot \mathbf{o}_i +  \mathbf{w}_2 \cdot \mathbf{a}_j +  \mathbf{w}_3 \cdot (\mathbf{o}_i \otimes \mathbf{a}_j) 
\end{equation}
where $\otimes$ denotes element-wise product and $\mathbf{w}_1, \mathbf{w}_2, \mathbf{w}_3$ are learnable vectors. The observation-contextualized vector for $j$-th action token is then 
\begin{align}
    \mathbf{ca}_j &= \mathbf{w}_5 \cdot \text{leakyRELU} (\mathbf{w}_4 \cdot [\mathbf{a}_j, \mathbf{c}_j, \mathbf{a}_j \otimes \mathbf{c}_j, \mathbf{q} \otimes \mathbf{c}_j]) \\
    \mathbf{c}_j &= \frac{ \sum_i \exp(\alpha_{ij}) \cdot \mathbf{o}_i}{ \sum_i \exp(\alpha_{ij})}, 
    \quad \mathbf{q} = \frac{ \sum_{j'} \exp(\max_i \alpha_{ij'}) \mathbf{a}_{j'}}{ \sum_{j'} \exp(\max_i \alpha_{ij'})}
\end{align}

We then average pool all $\mathbf{ca}_j$ to derive the action score $S(o, a)$:
\begin{equation}
    S(o, a) =  \mathbf{w}_6 \cdot  \frac{ 1}{n_a} \sum_{j \le n_a} \mathbf{ca}_j \in \mathbb{R}
\end{equation}
where $n_a$ is the number of tokens for action $a$. 

\subsection{RNN Baseline}

Our RNN baseline is inspired by \citet{guo2020interactive}, where we use the same attention layer as described above, but replace the Transformer text encoder with one-layer bi-directional Gated Recurrent Units (GRU)~\cite{Chung2014EmpiricalEO} of hidden dimension 512. Another difference is that we also add an cross attention between the instruction and action input word embeddings, as we hypothesize it might help option text matching. 

\section{\benchmark{} Experiment Details}
\label{appx:exp_setup}

\subsection{IL Training Details}
The training code for our IL models is adapted from \href{https://github.com/huggingface/transformers/blob/main/examples/pytorch/text-classification/run_glue_no_trainer.py}{Huggingface glue training example}, whose repository is licensed under Apache License 2.0. We use a training batch size of 1 with 32 gradient accumulation steps, a learning rate of $2 \times 10 ^ {-5}$, and 10 training epochs. The training takes around 2 hours on one RTX 2080 GPU with a GPU memory of around 10GB.

\subsection{RL Training Details}
We train the RL models using 4 parallel environments for $100,000$ training steps. The backprogation through time (BPTT) is taken at every 8 steps. We use an Adam optimizer with a learning rate of $10^{-5}$ (for Transformer models) or $5 \times 10^{-4}$ (for RNN models).

For RL models with the Transformer (BERT) architecture, it takes around 27 hours on one RTX 3090 GPU with a GPU memory of around 20GB. For RL models with the GRU architecture, it takes around 20 hours on one RTX 2080 GPU with a GPU memory of around 10GB.

To disentangle the effects of learning to search from choosing the right actions, we construct a \choice{} oracle that has access to the hidden reward function as well as hidden attributes and options underlying each product and instruction.\footnote{\color{black}A similar search oracle is also possible but harder to design since the search space is infinite. One possible oracle is to search for the underlying product name for each instruction, but that makes choice trivial as the underlying product is then almost always the first search result.} Given a search query, the %

{\color{black}
\subsection{Sampling vs. Top-1}
We show comparisons between using beam search vs. top-1 for both the search model and the choice model in Table~\ref{table:beam_vs_top1}.
During testing, the search model uses beam search to generate top-5 search queries. We randomly and uniformly sample from the top-5 queries to increase search diversity in case of multiple searches. We conduct experiments to instead always use the top-1 search, which shows slight performance improvement (see table below), and we will include the result in the paper.
The choice model has a fixed set of action candidates at each step (e.g. all available buttons), and we sample from the choice policy what action to take, as always taking the top action will lead to significantly detorior performances.
}
\begin{table}[t]
    \centering
    \footnotesize
\begin{tabular}{lll}
\toprule
 & Score & SR \\ 
\midrule
\color{black}{IL} & \color{black}{60.56 (1.94)} & \color{black}{29.00 (2.42)} \\
\color{black}{IL (top-1 search)} & \color{black}{61.96 (0.47)} & \color{black}{30.80 (0.72)} \\
\color{black}{IL (top-1 choice)} & \color{black}{45.10 (3.50)} & \color{black}{24.93 (3.14)} \\
\bottomrule
\end{tabular}
\vspace{3pt}
\caption{
\color{black}{Sampling vs. top-1.}
}
\label{table:beam_vs_top1}
\vspace{-10pt}
\end{table}

{\color{black}
\subsection{Image Ablation}
We train 3 trials with different random seeds for both the IL model and the ablated IL model without images, with performances over 500 test cases (\ref{table:image_ablation}).
Removing image only slightly hurts the overall performance, but significantly reduces the variance. This is reasonable as our current instruction and reward setups only use textual information, and we believe future efforts to incorporate visual information into the task setup will better challenge models’ visual understanding, and make pre-trained vision-language models such as CLIP more useful.
}
\begin{table}[t]
    \centering
    \footnotesize
\begin{tabular}{lll}
\toprule
 & Score & SR \\ 
\midrule
\color{black}{IL} & \color{black}{60.6 (1.94)} & \color{black}{29.0 (2.42)} \\
\color{black}{IL (w/o image)} & \color{black}{60.3 (0.47)} & \color{black}{28.4 (0.87)} \\
\bottomrule
\end{tabular}
\vspace{3pt}
\caption{
\color{black}{Image ablations. }
}
\label{table:image_ablation}
\vspace{-10pt}
\end{table}

\section{Sim-to-real Details}
\label{appx:sim2real}

\subsection{Sim-to-real Transfer Details}
\label{simRealTransfer}

To test how well our IL agent trained in \benchmark{} performs on \url{amazon.com} (\url{ebay.com} similarly), we wrote a series of scripts that generally achieve two steps - translate a real Amazon URL into our IL model's input (text observation, set of valid actions) and map the model's output back to a real Amazon URL. The following procedure is repeated until the IL model generates a "buy now" action:

\begin{itemize}
    \item Amazon URL $\rightarrow$ Amazon HTML $\rightarrow$ Amazon Page Information: Using ScraperAPI~\cite{scraperapi}, we first get the HTML source code for a given Amazon page, then extract information relevant to rendering the equivalent page in the \benchmark{} environment (e.g.\,title, price, options).
    \item Amazon Page Information $\rightarrow$ \benchmark{} HTML $\rightarrow$ Text Observation: Given the scraped information, we generate the corresponding \benchmark{} page in \html{} mode, then transform it into a \clean{} mode text observation.
    \item Amazon Page Information $\rightarrow$ Valid Action Set: From the scraped information, we determine what valid actions the model can take (i.e. $\search{\text{Red shoes}}$, $\click{\text{Size 9}}$). This logic is captured as a mapping of page type to permissible actions.
    \item Text Observation, Valid Action Set $\rightarrow$ IL Model $\rightarrow$ Amazon URL: Given the text observation and allowed of valid actions, the IL model produces an action. This action is then used to construct a corresponding Amazon URL via a set of mapping rules, and the loop is repeated. This continues until the model generates a "buy now" action, terminating the loop.
\end{itemize}

\subsection{Sim-to-real Transfer Results}

{\color{black}

The resulting numbers in Table~\ref{table:sim_real_stats} closely cohere to the reported numbers of \benchmark{} found in Figure~\ref{fig:main_results}, suggesting that the \benchmark{} has promise for developing grounded agents that can operate on real web environments. Between the two websites, transfer to Amazon is better than eBay as we note that (i) eBay has a larger product gap from \benchmark{}, e.g.\,some item categories like food are disallowed in eBay. (ii) the eBay search engine seems weaker, and would sometimes display no results for lengthy instructions. The following Table~\ref{table:transfer_example} is an example of a trajectory generated by the IL agent searching on the real Amazon website.

\begin{table}[!ht]
\begin{tabularx}{\textwidth}{X}
\toprule
{\centering\textbf{Instruction}}:
{I want to find \textcolor{mellowred}{white blackout} shades that are     \textcolor{mellowred}{66 inches in width and 66 inches in height}. They need to be \textcolor{mellowblue}{easy to install}..}
\\ \midrule 
search[\textcolor{mellowred}{white blackout} shades \textcolor{mellowred}{66 inches in width and 66 inches height}, \textcolor{mellowblue}{easy to install}]
click[item - Easy Up \& Down 100\% Blackout Pleated Window Shades Temporary Window Blinds 36in x 64in (Fits Window Width 18"-36")  2pcs-Pack Operating with Pull Cord  Easy Trimming \& Installing]
click[features]
click[back to search]
search[\textcolor{mellowred}{white blackout} shades that are \textcolor{mellowred}{66 inches in width and 66 inches height}]
click[item - Redi Shade Inc 1617201 Original Blackout Pleated Paper Shade Black 36” x 72” 6-Pack]
click[< prev]
click[Shade + Strips, \textcolor{mellowred}{White}]
click[buy]
\\ \bottomrule
\end{tabularx}
\vspace{3pt}
\caption{An example trajectory (showing only actions) from the IL agent on the real Amazon website. We omit instructions and some human actions for instruction and trim item names for readability. Red denotes options and blue denotes attributes.}
\label{table:transfer_example}
\end{table}
\vspace{-10pt}

It is evident that the exploratory behavior and patterns learned and exhibited by the agent within the \benchmark{} environment is not lost in this transfer. These results point to the opportunity for sim-to-real trained agents to transfer to other real-world web tasks despite the domain shift in both data (products) and dynamics (search engine) With that said, the gap between human and model performance also encourage us to look into expanding on the current limitations in our work regarding both the model and the \benchmark{} environment.
\section{Potential Societal Impacts and Limitations}

\benchmark{} is designed to minimize human efforts in data collection and processing, but there are still potential concerns regarding diversity, fairness, and representation. Developing RL agents that interact with the web also bear safety concerns, especially when transferring from simulation to real-world websites. We also discuss other limitations regarding the semantics of current task (instruction/reward).

\myparagraph{Diversity and representation in data collection.} We chose five common categories from \url{amazon.com} and scrape all products using all subcategories to minimize bias. However, our data is still biased toward the website country (USA) and website language (English), and may only represent a subset of all possible products that users potentially want to buy. Having this limitation in mind, the design of \benchmark{} allows the product data to be easily updated for different representations of real-world usage. 

\myparagraph{Bias in data processing.} Currently our attribute labeling is manually done and may be biased by the labeller's own experience (e.g.\,more knowledge toward product attributes like sports rather than makeup). An more automatic alternative would be to employ trained NLP models (e.g.\,relation extraction) to extract product attributes, which might be less biased than one labeller. Our reward design is general and could be updated to weight more toward attributes, options, price, etc. 

\myparagraph{Safety for developing web agents.}  Unlike recent work~\cite{nakano2021webgpt} that directly employs agents on the World Wide Web (WWW), \benchmark{} aims to provide a realistic simulation environment to train agents in a controllable and safe manner. In our preliminary sim-to-real experiments, the agent could only update the current webpage's url in two fixed and safe ways (i.e.\,search for results, open an item), and any form sending action (e.g.\,click options or buy) is held within the sim-to-real interface for later reward calculation. As a result, only navigation is done on the real-world website. For future deployment to real-world websites with more advanced functions, we believe a good specification of possible model behaviors is key to avoid harmful actions.

\myparagraph{Limitations in the current task.} Our current instructions are still limited by the attributes and options used. While attributes are simple and sometimes too generic (e.g.\,``easy to use''), the options might get too specific (e.g.\,``d17(dedicated right, back)''). Therefore, an agent might sometimes use a special option as cues to find the product, while ignoring other parts of the instruction. To better leverage images and texts (including reviews written by human users, which are not used in current work) of products for more semantic and challenging instructions is an important future direction from \benchmark{}.

\section*{Instruction for Human Trajectory Collection}
The following pages display the human trajectory collection document mentioned in \S\ref{appx:traj}.



\section*{Supplementary material}

# The WebShop Task

Thank you for taking part in this project! In this task, you need to buy a designated product given an instruction on our Amazon Shopping Game site. You will get a score in the end indicating how close you are. Please try to score as high as you can.

If you find in some cases the scoring seems weird/unfair, please reach out. We will look into the cases.

Please read the following instructions carefully before you start.

## Instructions

**1**) Go to the home page. The instruction will immediately show up on the landing page.

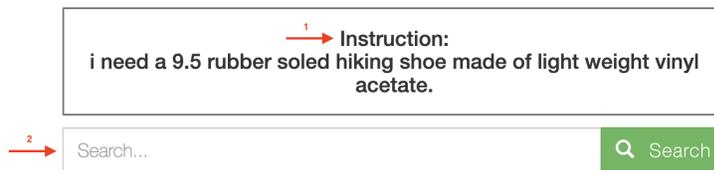

**2**) Given this instruction, please write a search query that would produce search results matching the description.

Please *do not* copy-paste the entire instruction. We encourage you come up with more targeted queries, see the result, and search again if needed.
Example:
- `Instruction`:
  I need a 9.5 rubber soled hiking shoe made of lightweight vinyl acetate.

- `Bad query`: (copy pasting)
  9.5 rubber soled hiking shoe made of lightweight vinyl acetate

- `Ideal query`: (1st attempt)
    rubber soled hiking shoe vinyl acetate (say the results are not great)
  - `Ideal query`: (2nd attempt)
    hiking shoe lightweight vinyl acetate  (the results are better)
  - `Ideal query`: (3nd attempt)
    lightweight climbing shoe vinyl        (gives promising results)

Essentially, you need to hack the search engine a little bit.

Note that our search engine is limited. Tricks that work on Google Search such as adding quotation marks around the query won't work.

Click **Search** after filling out the search bar like below.

### Amazon Shopping Game

> Instruction:
> i need a 9.5 rubber soled hiking shoe made of light weight vinyl acetate.

`rubber-soled lightweight vinyl acetate hiking shoes size 9.5`  🔍 Search

**3**) Upon clicking **Search**, you will be sent to a page of results. The below screenshot is an example of the results displayed from the example query in Step 2. Each page shows up to 10 results. Click the **Next** button to see more results.

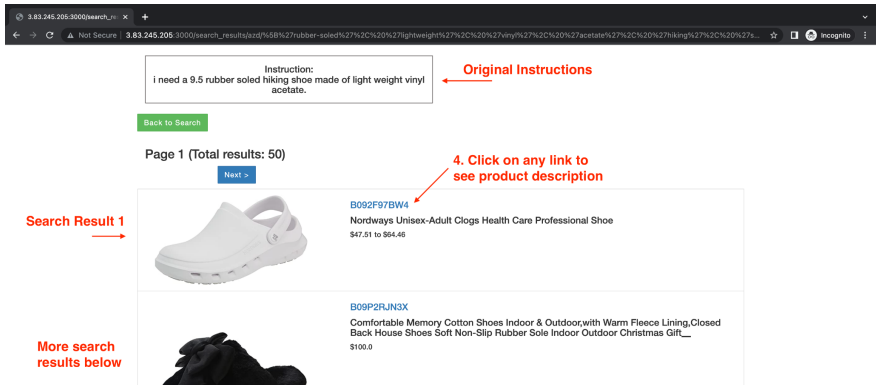

**4**) Click on any of the blue product title text (i.e. *"B092F97B24"* in above screenshot) to see a product detail page, like the below.

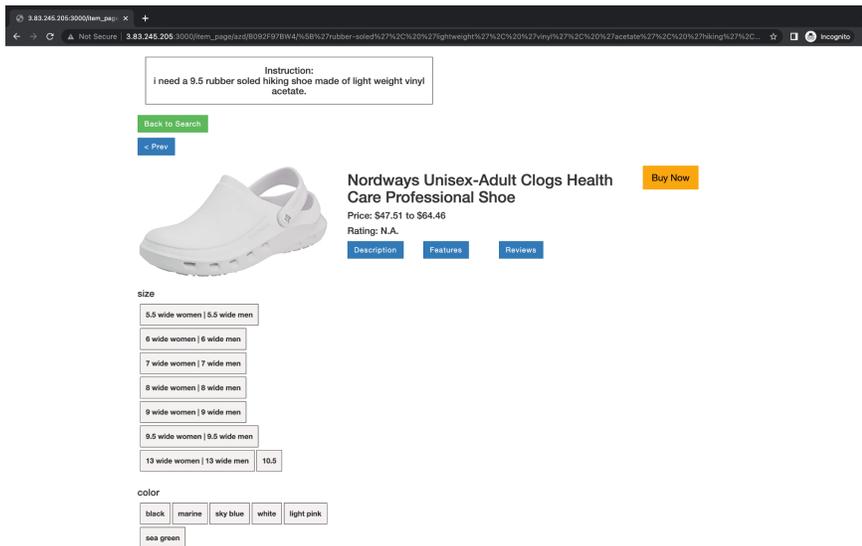

# Guidelines for searching for matching products:

- Some pages have **Options** (i.e. *Size*, *Color* in above screenshot). If the instructions contain such information, please select the corresponding options (even if the title / features / desc. / reviews may already contain such info). In most cases, if you find the options verbatim as in the instruction, you've likely found the right product.
- Do **not** use the product image to determine whether the instruction's information matches the product.

  An example:
  - Given instruction: "Find me a pair of ankle socks that are blue and size 11"

Between this product…  And this product…

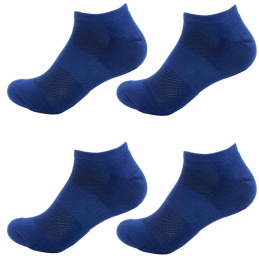 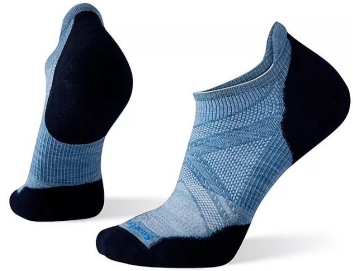

**TITLE**: Ankle socks for casual wear, sports, and leisure. Pack of 4, 8, or 12

**DESC**: 100% made in the USA. These socks are good for any occasion.
**FEATURES**: Made with cotton, breathable fabric. Machine washable okay.

**OPTIONS**:
- Sizes: 8, 9, 10, 11, 12
- Color: red, green, black, white, blue

**TITLE**: Kirkland athletic socks with rubber soles and heels. Easy slip on

**DESC**: Costco wholesale socks, limited stock.
**FEATURES**: Polyester and Rayon fabric. Guaranteed long lasting or your money back.

**OPTIONS**: None

The left hand is a **better match** because the product's title, features, description, and options reflect the instruction's information.

While the right hand product appears to be a pair of blue ankle socks, because this information is not reflected in the text, we do **not** consider this a match.

> Therefore, feel free to use the product image as a reference when looking for matches, but keep in mind that the experiment we're running accounts for a text's

**5**) Decide whether the product is a match

> A **match** should
> - Contain all of the instruction's information in the product detail page's text (i.e. title, description, feature, options)
> - Have options (if they exist), which correspond to the product info, be selected.
>
> A match does **not** account for
> - The product image

- *You think it is a match!* → Click the <mark>Buy Now</mark> button on the product detail page
- *You think it is not a match OR another product might be a better match*…
    - Click on the Back button to go to the original list of search results (page **3**). From here, repeat steps **3-4** until you find a product that matches best.
    - Click on the Back to Search button. This will take you back to the search bar page (page **2**). If you feel none of the results are good matches, try another search query.

**6**) Once you clicked <mark>Buy Now</mark>, you will see your score (won't be used to decide the pay), and a code you need to paste in the MTurk interface. And you're done!

## Tips

Patterns that often result in HIGH scores:
- Refine search queries until promising products show up
- Explore different product pages (go to next page if needed) to see if options and different aspects are covered
- Make sure all aspects in the instructions are covered by either the title, description page, or the feature page.
- Make sure all options are found almost verbatim in the product page

Patterns that often result in LOW scores:

- Low effort copy-paste the entire instructions as the search query
- Always click the first item without checking if the aspects in the instructions are covered
- Click items that obviously don't have any option matches



\end{document}